\documentclass[10pt,twocolumn,letterpaper]{article}

\usepackage[pagenumbers]{cvpr} %

\usepackage{graphicx, amsmath, amssymb, caption, multirow, overpic, textpos}
\usepackage[table]{xcolor}
\usepackage[british, english, american]{babel}
\usepackage{booktabs}
\usepackage{times}
\usepackage{epsfig}
\usepackage{tabulary}
\usepackage{soul}
\usepackage{xspace}\xspace
\usepackage{adjustbox}
\usepackage{array}
\usepackage[pagebackref=false, breaklinks=true, letterpaper=true, colorlinks,
            citecolor=citecolor, linkcolor=linkcolor, bookmarks=false]{hyperref}
\definecolor{citecolor}{HTML}{0071BC}
\definecolor{linkcolor}{HTML}{ED1C24}
\newcommand{\cmark}{\checkmark} %

\usepackage{dblfloatfix}
\usepackage{transparent}

\usepackage{algorithm}
\usepackage{listings}
\usepackage{scalerel}

\usepackage{cuted}
\usepackage{capt-of}

\newlength\savewidth\newcommand\shline{\noalign{\global\savewidth\arrayrulewidth
  \global\arrayrulewidth 1pt}\hline\noalign{\global\arrayrulewidth\savewidth}}
\newcommand{\tablestyle}[2]{\setlength{\tabcolsep}{#1}\renewcommand{\arraystretch}{#2}\centering\footnotesize}
\renewcommand{\paragraph}[1]{\vspace{1.25mm}\noindent\textbf{#1}}
\newcommand\blfootnote[1]{\begingroup\renewcommand\thefootnote{}\footnote{#1}\addtocounter{footnote}{-1}\endgroup}

\newcolumntype{x}[1]{>{\centering\arraybackslash}p{#1pt}}
\newcolumntype{y}[1]{>{\raggedright\arraybackslash}p{#1pt}}
\newcolumntype{z}[1]{>{\raggedleft\arraybackslash}p{#1pt}}

\newcommand{\app}{\raise.17ex\hbox{$\scriptstyle\sim$}}

\definecolor{baselinecolor}{gray}{.9}
\definecolor{deemph}{gray}{0.6}
\newcommand{\gc}[1]{\textcolor{deemph}{#1}}
\newcommand{\gr}{\rowcolor[gray]{.95}}

\newcommand{\baseline}[1]{\cellcolor{baselinecolor}{#1}}

\usepackage{etoolbox}
\makeatletter
\AfterEndEnvironment{algorithm}{\let\@algcomment\relax}
\AtEndEnvironment{algorithm}{\kern2pt\hrule\relax\vskip3pt\@algcomment}
\let\@algcomment\relax
\newcommand\algcomment[1]{\def\@algcomment{\footnotesize#1}}
\renewcommand\fs@ruled{\def\@fs@cfont{\bfseries}\let\@fs@capt\floatc@ruled
  \def\@fs@pre{\hrule height.8pt depth0pt \kern2pt}%
  \def\@fs@post{}%
  \def\@fs@mid{\kern2pt\hrule\kern2pt}%
  \let\@fs@iftopcapt\iftrue}
\makeatother

\newcommand{\convmae}{FCMAE\xspace}

\usepackage[capitalize]{cleveref}
\crefname{section}{Sec.}{Secs.}
\Crefname{section}{Section}{Sections}
\Crefname{table}{Table}{Tables}
\crefname{table}{Tab.}{Tabs.}

\def\cvprPaperID{0000} %
\def\confName{CVPR}
\def\confYear{2023}

\begin{document}

\title{ConvNeXt V2: Co-designing and Scaling ConvNets with Masked Autoencoders}

\author{
Sanghyun Woo\textsuperscript{1*} \quad 
Shoubhik Debnath\textsuperscript{2} \quad
Ronghang Hu\textsuperscript{2} \quad \\
Xinlei Chen\textsuperscript{2} \quad 
Zhuang Liu\textsuperscript{2} \quad
In So Kweon\textsuperscript{1} \quad
Saining Xie$^{3\dagger}$\\
 \textsuperscript{1}KAIST
 \qquad \textsuperscript{2} Meta AI, FAIR
\qquad \textsuperscript{3}New York University
}
\maketitle

\blfootnote{\textsuperscript{*} Work done during an internship at FAIR.}
\blfootnote{$^\dagger$ Corresponding author.}

\begin{abstract}
Driven by improved architectures and better representation learning frameworks, the field of visual recognition has enjoyed rapid modernization and performance boost in the early 2020s. For example, modern ConvNets, represented by ConvNeXt~\cite{liu2022convnet}, have demonstrated strong performance in various scenarios. While these models were originally designed for supervised learning with ImageNet labels, they can also potentially benefit from self-supervised learning techniques such as masked autoencoders (MAE)~\cite{he2022masked}. However, we found that simply combining these two approaches leads to subpar performance. In this paper, we propose a fully convolutional masked autoencoder framework and a new Global Response Normalization (GRN) layer that can be added to the ConvNeXt architecture to enhance inter-channel feature competition. This co-design of self-supervised learning techniques and architectural improvement results in a new model family called ConvNeXt V2, which significantly improves the performance of pure ConvNets on various recognition benchmarks, including ImageNet classification, COCO detection, and ADE20K segmentation. We also provide pre-trained ConvNeXt V2 models of various sizes, ranging from an efficient 3.7M-parameter Atto model with 76.7\% top-1 accuracy on ImageNet, to a 650M Huge model that achieves a state-of-the-art 88.9\% accuracy using only public training data.
\end{abstract}

\begin{textblock*}{.8\textwidth}[.5,0](0.53\textwidth, -.74\textwidth)
\centering
{\hspace{-6ex} \small Code: \url{https://github.com/facebookresearch/ConvNeXt-V2}}
\end{textblock*}

\section{Introduction}

\begin{figure}\centering
\includegraphics[width=0.48\textwidth]{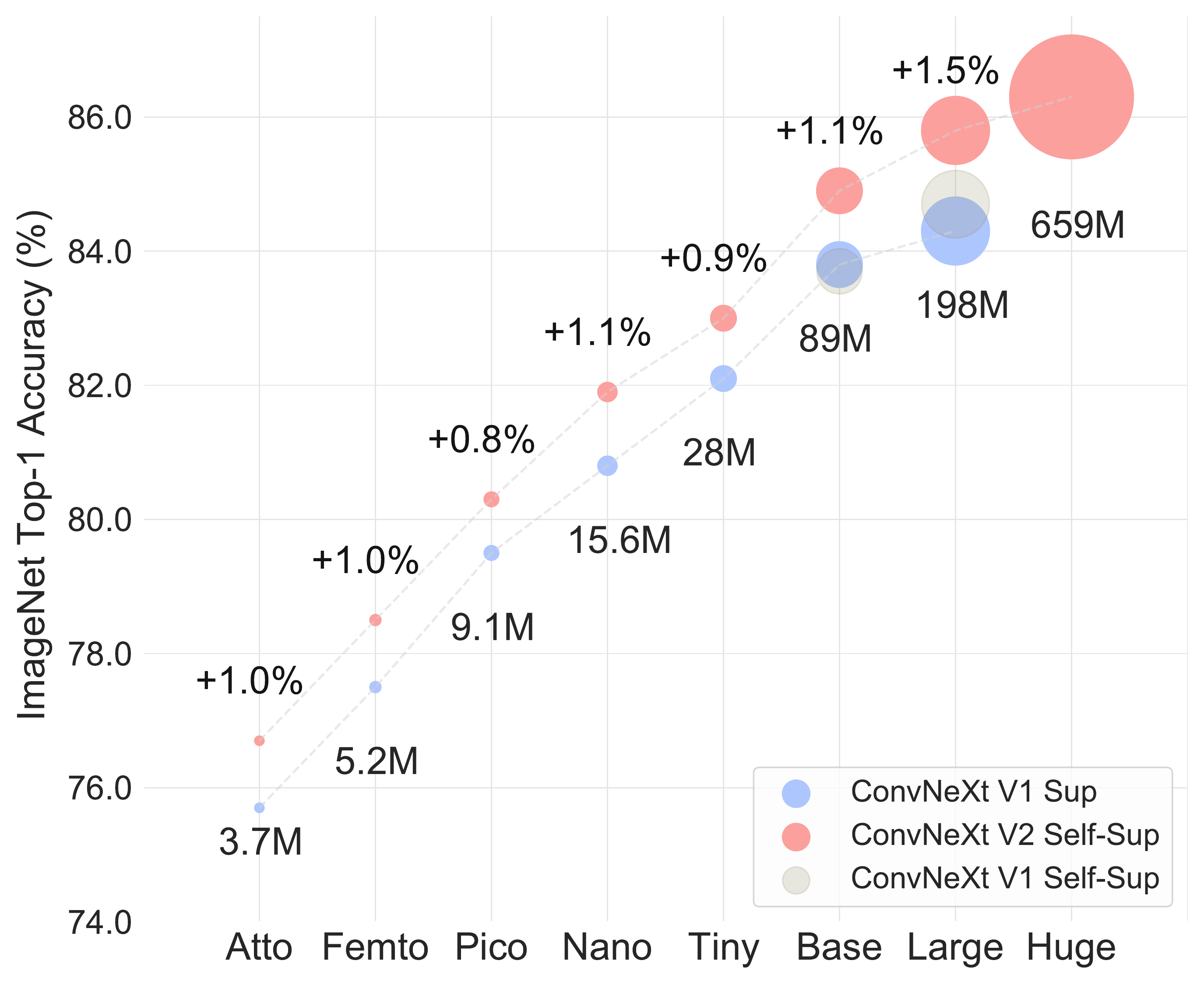}
\vspace{-1.5em}
\captionof{figure}{
\textbf{ConvNeXt V2 model scaling}. 
The ConvNeXt V2 model, which has been pre-trained using our fully convolutional masked autoencoder framework, performs significantly better than the previous version across a wide range of model sizes. \label{fig:teaser}}
\vspace{-1em}
\end{figure}

Building on research breakthroughs in earlier decades \cite{lecun1998gradient, Russakovsky2015, Krizhevsky2012, He2015, vaswani2017attention}, the field of visual recognition has ushered in a new era of large-scale visual representation learning. Pre-trained, large-scale vision models have become essential tools for feature learning and enabling a wide range of vision applications. The performance of a visual representation learning system is largely influenced by three main factors: the neural network architecture chosen, the method used for training the network, and the data used for training. In the field of visual recognition, progress in each of these areas contributes to overall improvements in performance.

Innovation in neural network architecture design has consistently played a major role in the field of representation learning. Convolutional neural network architectures (ConvNets)~\cite{lecun1998gradient, Krizhevsky2012, He2015} have had a significant impact on computer vision research by allowing for the use of generic feature learning methods for a variety of visual recognition tasks~\cite{Girshick2014,He2017}, rather than relying on manual feature engineering. In recent years, the transformer architecture~\cite{vaswani2017attention}, originally developed for natural language processing, has also gained popularity due to its strong scaling behavior with respect to model and dataset size~\cite{Dosovitskiy2021}. More recently, ConvNeXt~\cite{liu2022convnet} architecture has modernized traditional ConvNets and demonstrated that pure convolutional models could also be scalable architectures. However, the most common method for exploring the design space for neural network architectures is still through benchmarking \emph{supervised learning} performance on ImageNet.

In a separate line of research, the focus of visual representation learning has been shifting from supervised learning with labels to self-supervised pre-training with pre-text objectives. Among many different self-supervised algorithms, masked autoencoders (MAE)~\cite{he2022masked} have recently brought success in masked language modeling to the vision domain and quickly become a popular approach for visual representation learning. 
However, a common practice in self-supervised learning is to use a \emph{predetermined} architecture designed for supervised learning, and assume the design is fixed. For instance, MAE was developed using the vision transformer~\cite{Dosovitskiy2021}  architecture.

It is possible to combine the design elements of architectures and self-supervised learning frameworks, but doing so may present challenges when using ConvNeXt with masked autoencoders. One issue is that MAE has a specific encode-decoder design that is optimized for the sequence processing capabilities of transformers, which allows the compute-heavy encoder to focus on visible patches and thus reduce the pre-training cost. This design may not be compatible with standard ConvNets, which use dense sliding windows. Additionally, if the relationship between the architecture and the training objective is not taken into consideration, it may be unclear whether optimal performance can be achieved. In fact, previous research has shown that training ConvNets with mask-based self-supervised learning can be difficult~\cite{jing2022masked}, and empirical evidence suggests that transformers and ConvNets may have different feature learning behaviors that can affect representation quality.

To this end, we propose to \emph{co-design} the network architecture and the masked autoencoder under the same framework, with the aim of making mask-based self-supervised learning effective for ConvNeXt models and achieving results similar to those obtained using transformers.

In designing the masked autoencoder, we treat the masked input as a set of sparse patches and use sparse convolutions~\cite{graham2017submanifold} to process only the visible parts. The idea is inspired by the use of sparse convolutions in processing large-scale 3D point clouds~\cite{choy20194d, xie2020pointcontrast}. In practice, we can implement ConvNeXt with sparse convolutions, and at fine-tuning, the weights are converted back to standard, dense layers without requiring special handling. To further improve the pre-training efficiency, we replace the transformer decoder with a single ConvNeXt block, making the entire design fully convolutional. We have observed mixed results with these changes: the learned features are useful and improve upon the baseline results, but the fine-tuning performance is still not as good as the transformer-based model.

We then conduct a feature space analysis of different training configurations for ConvNeXt. We identify a potential issue of feature collapse at the MLP layer when training ConvNeXt directly on masked input. To address this issue, we propose adding a Global Response Normalization layer to enhance inter-channel feature competition. This change is most effective when the model is pre-trained with masked autoencoders, suggesting that reusing a fixed architecture design from supervised learning may be suboptimal.

In summary, we introduce ConvNeXt V2 which demonstrates improved performance when used in conjunction with masked autoencoders. We have found that this model significantly improves the performance of pure ConvNets across various downstream tasks, including ImageNet classification~\cite{Russakovsky2015}, COCO object detection~\cite{Lin2014} and ADE20K segmentation~\cite{Zhou2014}. 
The ConvNeXt V2 models can be used in a variety of compute regimes and includes models of varying complexity: from an efficient 3.7M-parameter \emph{Atto} model that achieves 76.7\% top-1 accuracy on ImageNet, to a 650M \emph{Huge} model that reaches a state-of-the-art 88.9\% accuracy when using IN-22K labels.

\section{Related Work}
\paragraph{ConvNets.} The design of ConvNets, which were first introduced in the 1980s~\cite{LeCun1989} and trained using back-propagation, has undergone numerous improvements in terms of optimization, accuracy, and efficiency over the years~\cite{Krizhevsky2012, Szegedy2015, He2016, Xie2017, Huang2017, Howard2017, Tan2019efficientnet, Radosavovic2020designing}. These innovations have mainly been discovered through the use of supervised training on the ImageNet dataset. In recent years, some efforts have been made to perform architecture search using self-supervised pre-text tasks such as rotation prediction and colorization, as in the case of UnNAS~\cite{liu2020labels}. Recently, ConvNeXt~\cite{liu2022convnet} conducted a comprehensive review of the design space and demonstrated pure ConvNets can be as scalable as the vision transformers~\cite{Dosovitskiy2021, Liu2021swin}, which have become the dominant architecture in many applications. ConvNeXt has particularly excelled in scenarios requiring lower complexity~\cite{capelle2022finding, rw2022which, rw2019timm}. Our ConvNeXt V2 model, which is powered by self-supervised learning, provides a simple way to upgrade existing models and achieve a significant boost in performance across a wide range of use cases.

\paragraph{Masked Autoencoders.} 
Masked image modeling, represented by masked autoencoders~\cite{he2022masked}, is one of the latest self-supervised learning strategies. As a neural network pre-training framework, masked autoencoders have shown a broad impact on visual recognition. However, original masked autoencoders are not directly applicable to ConvNets due to their asymmetric encoder-decoder design. Alternative frameworks such as~\cite{Bao2021, xie2022simmim} have attempted to adapt the approach for use with ConvNets, but with mixed results. MCMAE~\cite{gao2022mcmae} uses a few convolutional blocks as input tokenizers.  To the best of our knowledge, there are no pre-trained models that show self-supervised learning can improve upon the best ConvNeXt supervised results.

\section{Fully Convolutional Masked Autoencoder}

\begin{figure}[t]
\centering
\vstretch{1.}{\includegraphics[width=0.48\textwidth]{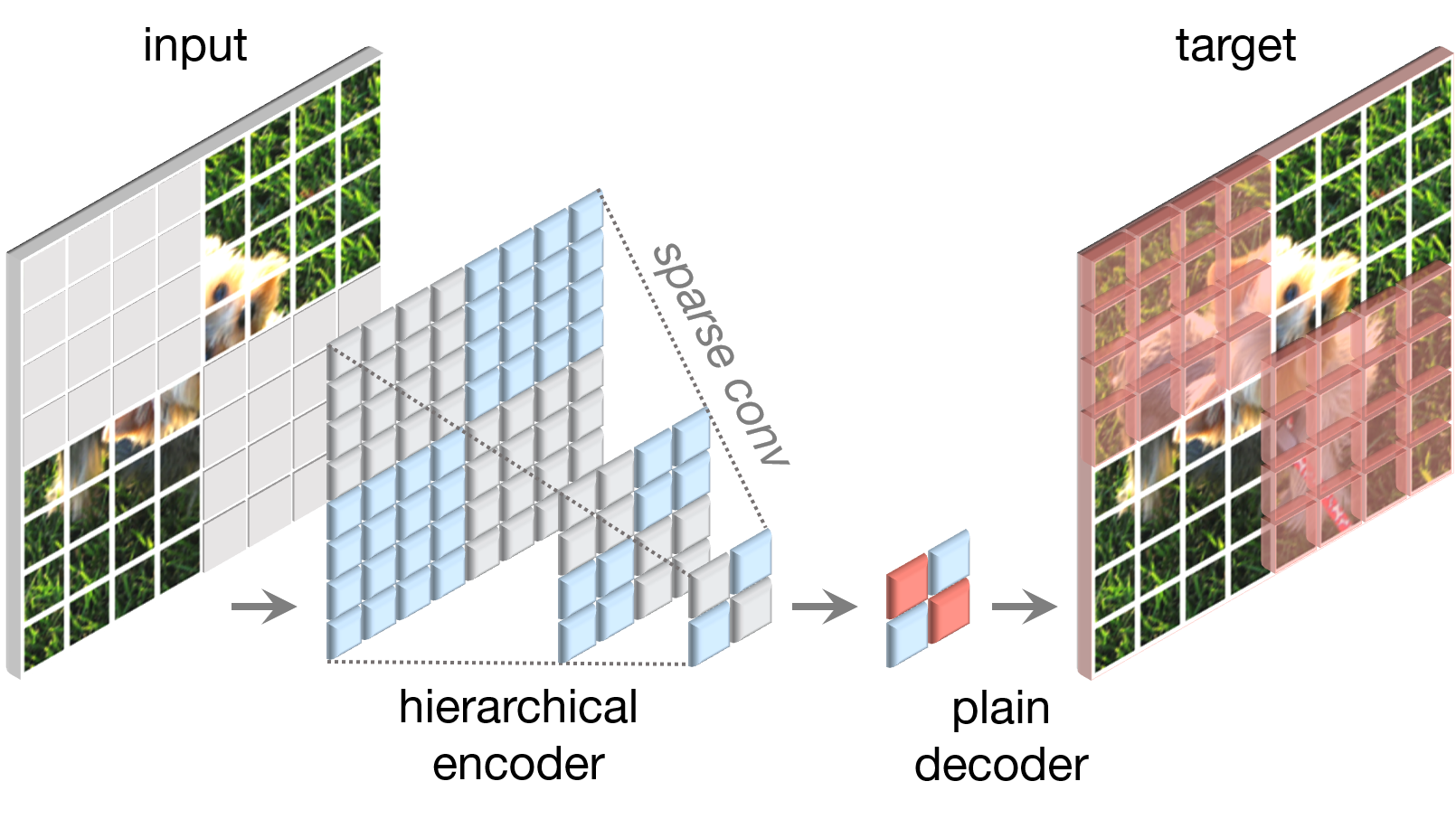}}
\vspace{-1em}
\caption{
\textbf{Our \convmae framework.}
We introduce a fully convolutional masked autoencoder (FCMAE). It consists of a sparse convolution-based ConvNeXt encoder and a lightweight ConvNeXt block decoder. Overall, the architecture of our autoencoder is asymmetric. The encoder processes only the visible pixels, and the decoder reconstructs the image using the encoded pixels and mask tokens. The loss is calculated only on the masked region.
}
\label{fig:convmae}
\vspace{-.5em}
\end{figure}

Our approach is conceptually simple and runs in a fully convolutional manner. The learning signals are generated by randomly \emph{masking} the raw input visuals with a high masking ratio and letting the model \emph{predict} the missing parts given the remaining context.
Our framework is illustrated in Figure~\ref{fig:convmae}, and we will now describe its main components in more detail.

\paragraph{Masking.} We use a random masking strategy with a masking ratio of 0.6. As the convolutional model has a hierarchical design, where the features are downsampled in different stages, the mask is generated in the last stage and upsampled recursively up to the finest resolution. To implement this in practice, we randomly remove 60\% of the $32\times32$ patches from the original input image. We use minimal data augmentation, only including random resized cropping.

\paragraph{Encoder design.}  We use ConvNeXt~\cite{liu2022convnet} model as the encoder in our approach.
One challenge in making masked image modeling effective is preventing the model from learning shortcuts that allow it to copy and paste information from the masked regions. This is relatively easy to prevent in transformer-based models, which can leave the visible patches as the only input to the encoder. However, it is more difficult to achieve this with ConvNets, as the 2D image structure must be preserved. 
While naive solutions involve introducing learnable masked tokens in the input side~\cite{Bao2021,xie2022simmim}, these approaches decrease the efficiency of pre-training and result in a train and test time inconsistency, as there are no mask tokens at test time. This becomes especially problematic when the masking ratio is high.

To tackle this issue, our new insight is to view the masked image from a ``{sparse data perspective}'', which was inspired by learning on sparse point clouds in 3D tasks~\cite{choy20194d,xie2020pointcontrast}.
Our key observation is that the masked image can be represented as a 2D sparse array of pixels. Based on this insight, it is natural to incorporate sparse convolution into our framework to facilitate pre-training of the masked autoencoder. In practice, during pre-training, we propose to convert the standard convolution layer in the encoder with the submanifold sparse convolution, which enables the model to operate \emph{only} on the visible data points~\cite{graham2017submanifold,graham20183d,choy20194d}. 
We note that the sparse convolution layers can be converted back to standard convolution at the fine-tuning stage without requiring additional handling. As an alternative, it is also possible to apply a binary masking operation before and after the dense convolution operation. This operation has numerically the same effect as sparse convolutions, is theoretically more computationally intensive, but can be more friendly on AI accelerators like TPU.

\begin{table*}[t!]
\vspace{-.2em}
\centering
\subfloat[
\textbf{Decoder design}.
\label{tab:decoder_design} A simple convolutional block outperforms more complex decoder designs.
]{
\begin{minipage}{0.35\linewidth}{\begin{center}
\tablestyle{1pt}{1.05}
\begin{tabular}{y{75}x{32}x{32}x{32}}
dec. type & ft & hours & speedup\\
\shline
{UNet {w/}  skip}               & \textbf{83.7}                    &  12.9 & {-}         \\
{UNet {w/o} skip}               & 83.5                             &  12.9 & {-}         \\
{Transformer~\cite{he2022masked}}   & 83.4                             &  8.5  & 1.5$\times$ \\
{ConvNeXt block}                           & \baseline{\textbf{83.7}}         &  \baseline{\textbf{7.7}}  & \baseline{\textbf{1.7$\times$}} \\

\multicolumn{4}{c}{~}\\
\end{tabular}
\end{center}}\end{minipage}
}
\hspace{1em}
\subfloat[
\textbf{Decoder depth}. A single block yields competitive fine-tuning performance.
\label{tab:decoder_depth}
]{
\centering
\begin{minipage}{0.26\linewidth}{\begin{center}
\tablestyle{4pt}{1.05}
\begin{tabular}{x{24}x{32}}
blocks & ft  \\
\shline
1 & \baseline{\textbf{83.7}} \\
2 & 83.5 \\
4 & \textbf{83.7} \\
8 & 83.6 \\
12 & 83.3 \\
\end{tabular}
\end{center}}\end{minipage}
}
\hspace{1em}
\subfloat[
\textbf{Decoder width}. A decoder width of 256 or 512 achieves the best performance.
\label{tab:decoder_width}
]{
\begin{minipage}{0.26\linewidth}{\begin{center}
\tablestyle{4pt}{1.05}
\begin{tabular}{x{24}x{32}}
dim & ft \\
\shline
128 & 83.5 \\
256 & \textbf{83.7}  \\
512 & \baseline{\textbf{83.7}} \\
768 & 83.6  \\
1024 & 83.5 \\
\end{tabular}
\end{center}}\end{minipage}
}
\caption{\textbf{MAE decoder ablation experiments} with ConvNeXt-Base on ImageNet-1K. We report fine-tuning (ft) accuracy (\%). 
The pre-training schedule is 800 epochs.
In the decoder design exploration, the wall-clock time is benchmarked on a 256-core TPU-v3 pod using JAX. 
The speedup is relative to the UNet decoder baseline.
Our final design choices employed in the paper are marked in \colorbox{baselinecolor}{gray}.}
\label{tab:dec_ablations} 
\end{table*}

\begin{figure*}[t]
\centering
\includegraphics[width=0.99\linewidth]{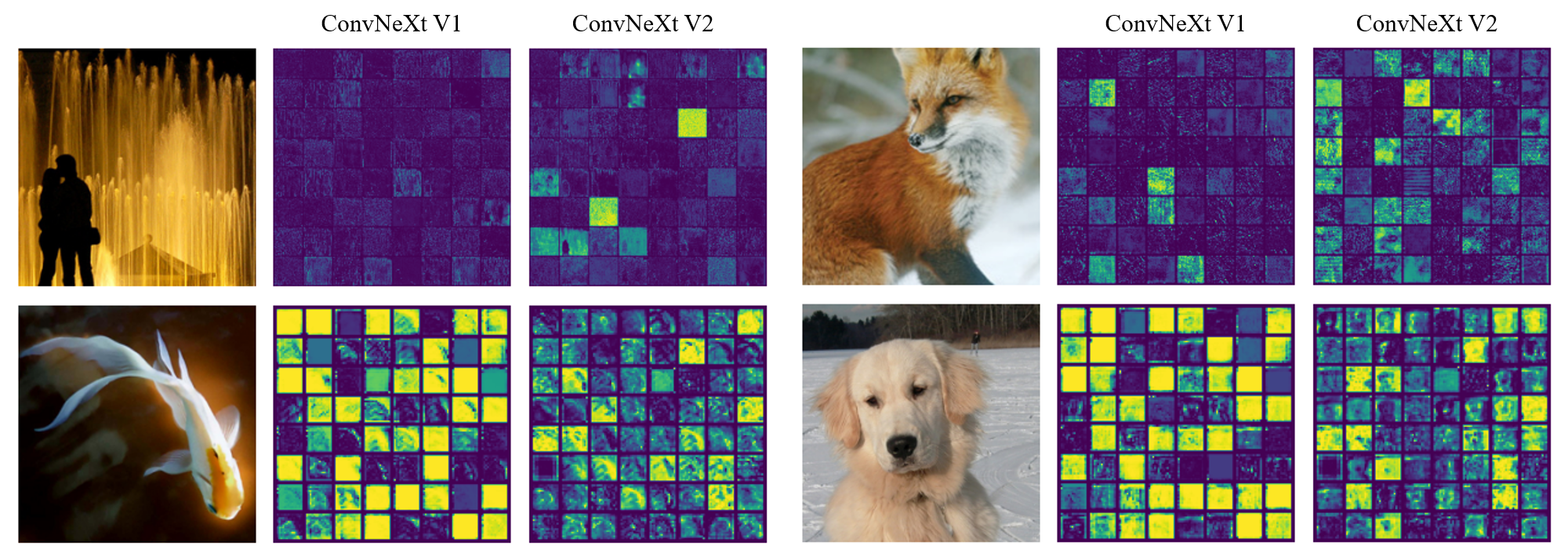}
\caption{
\textbf{Feature activation visualization.} We visualize the activation map for each feature channel in small squares. For clarity, we display 64 channels in each visualization. The ConvNeXt V1 model suffers from a feature collapse issue, which is characterized by the presence of redundant activations (dead or saturated neurons) across channels. To fix this problem, we introduce a new method to promote feature diversity during training: the global response normalization (GRN) layer. This technique is applied to high-dimensional features in every block, leading to the development of the ConvNeXt V2 architecture.
}
\label{fig:feat_act}
\end{figure*}

\paragraph{Decoder design.} 
We use a lightweight, plain ConvNeXt block as the decoder. This forms an asymmetric encoder-decoder architecture overall, as the encoder is heavier and has a hierarchy. We also considered more complex decoders such as hierarchical decoders~\cite{ronneberger2015u,lin2017feature} or transformers~\cite{Dosovitskiy2021,he2022masked}, but the simpler single ConvNeXt block decoder performed well in terms of fine-tuning accuracy and reduced pre-training time considerably, demonstrated in Table~\ref{tab:dec_ablations}. We set the dimension of the decoder to 512.

\paragraph{Reconstruction target.} We compute the mean squared error (MSE) between the reconstructed and target images. Similar to MAE~\cite{he2022masked}, the target is a patch-wise normalized image of the original input, and the loss is applied only on the masked patches.

\paragraph{\convmae.} 
We now present a \textbf{F}ully \textbf{C}onvolutional \textbf{M}asked \textbf{A}uto\textbf{E}ncoder (FCMAE) by combining the proposals described above. 
To evaluate the effectiveness of this framework, we use the ConvNeXt-Base model as the encoder and conduct a series of ablation studies. Throughout the paper, we focus on the end-to-end fine-tuning performance becuase of its practical relevance in transfer learning, and use that to assess the quality of the learned representation.

We pre-train and fine-tune using the ImageNet-1K (IN-1K) dataset for 800 and 100 epochs, respectively, and report the top-1 IN-1K validation accuracy for a single 224×224 center crop. Additional details about the experimental setup can be found in the appendix.

To understand the impact of using sparse convolution in our FCMAE framework, we first investigate how it affects the quality of the learned representation during masked image pre-training. Our empirical findings show that it is essential to prevent information leakage from the masked region in order to achieve good results.

\begin{center}\vspace{-.2em}
\tablestyle{4pt}{1.05}
\begin{tabular}{x{60}x{60}}
{w/o} Sparse conv. & w/ Sparse conv. \\
\shline
79.3 & 83.7
\end{tabular}\vspace{-.2em}
\end{center}

Next, we compare our self-supervised approach to supervised learning.
Specifically, we obtain two baseline experimental results: the supervised 100 epoch baseline using the same recipe and the 300 epoch supervised training baseline provided in the original ConvNeXt paper~\cite{liu2022convnet}. 
We find that our \convmae pre-training provides better initialization than the random baseline (\ie, 82.7 $\rightarrow$ 83.7), but it still needs to catch up to the best performance obtained in the original supervised setup.

\begin{center}\vspace{-.2em}
\tablestyle{4pt}{1.05}
\begin{tabular}{x{60}x{60}x{60}}
Sup, 100ep & Sup, 300ep.~\cite{liu2022convnet} & \convmae \\
\shline
82.7 & 83.8 & 83.7
\end{tabular}\vspace{-.2em}
\end{center}

This is in contrast to the recent success of masked image modeling using transformer-based models~\cite{Bao2021,he2022masked,xie2022simmim}, where the pre-trained models significantly outperform the supervised counterparts. This motivates us to investigate the unique challenges faced by the ConvNeXt encoder during masked autoencoder pre-training, which we discuss next.

\section{Global Response Normalization}
In this section, we introduce a new Global Response Normalization (GRN) technique to make \convmae pre-training more effective in conjunction with the ConvNeXt architecture. We first motivate our approach through both qualitative and quantitative feature analyses. 

\paragraph{Feature collapse.}
To gain more insight into the learning behavior, we first perform qualitative analysis in the feature space. We visualize the activations of a \convmae pre-trained ConvNeXt-Base model and notice an intriguing ``feature collapse'' phenomenon: there are many dead or saturated feature maps and the activation becomes redundant across channels. We show some of the visualizations in Figure \ref{fig:feat_act}. This behavior was mainly observed in the dimension-expansion MLP layers in a ConvNeXt block~\cite{liu2022convnet}.

\begin{figure}[t]
\centering\hspace{-5mm}
{\includegraphics[width=0.5\textwidth]{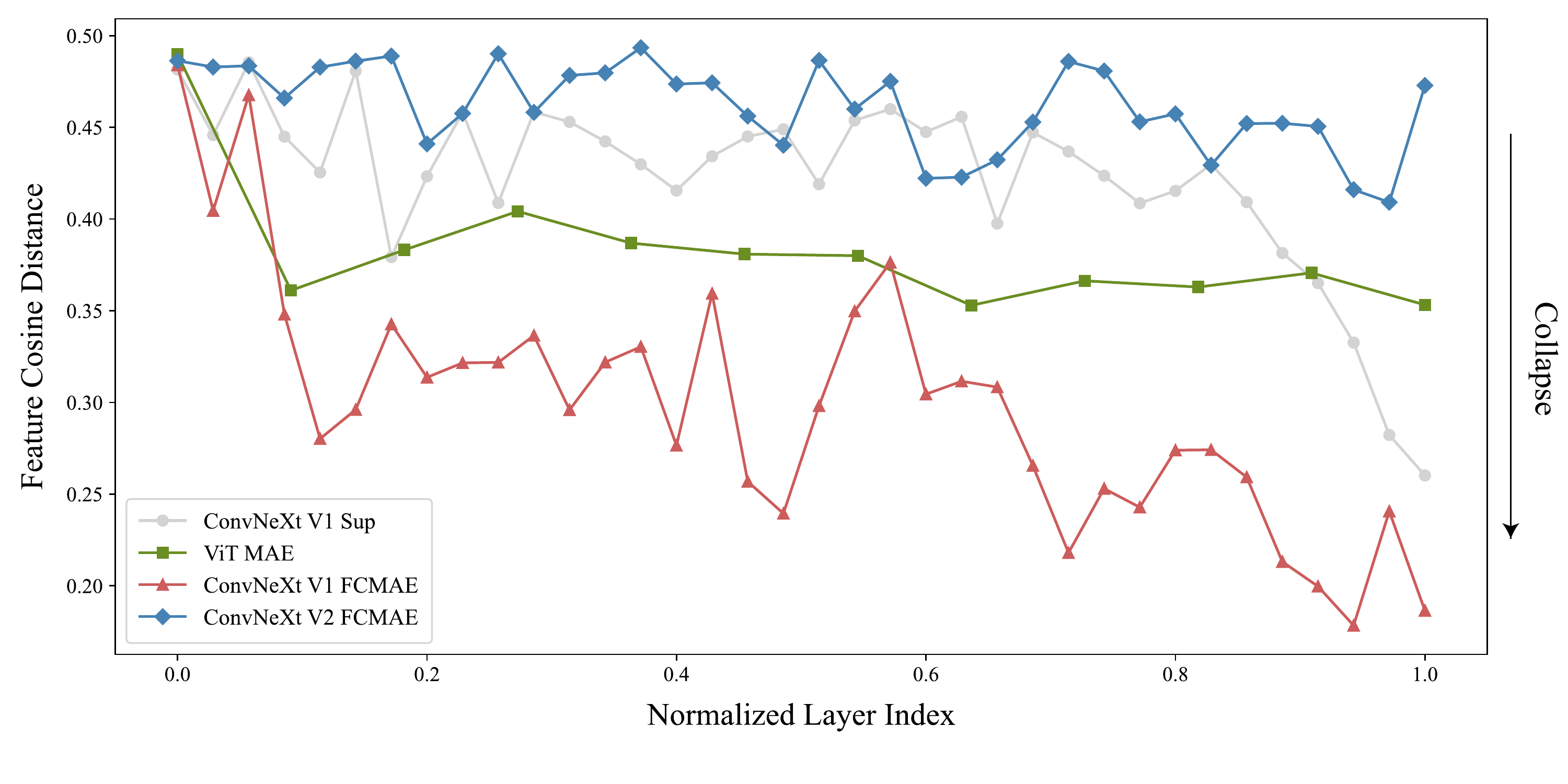}}
\vspace{-2em}
\caption{
\textbf{Feature cosine distance analysis}. As the number of total layers varies for different architectures, we plot the distance values against the normalized layer indexes. We observe that the ConvNeXt V1 \convmae pre-trained model exhibits severe feature collapse behavior. The supervised model also shows a reduction in feature diversity, but only in the final layers. This decrease in diversity in the supervised model is likely due to the use of the cross-entropy loss, which encourages the model to focus on class-discriminative features while suppressing the others.} 
\label{fig:feat_cos_dist}
\end{figure}

\paragraph{Feature cosine distance analysis.}
To further validate our observation quantitatively, we perform a feature cosine distance analysis. Given an activation tensor $X \in R^{H\times W\times C}$, $X_{i} \in R^{H \times W}$ is the feature map of the $i$-th channel. 
We reshape it as a $HW$ dimensional vector and compute the average pair-wise cosine distance across the channels by $\frac{1}{ C^{2}} \sum_{i}^{ C} \sum_{j}^{ C}  \frac{1-{cos}( X_{i}, X_{j})}{2}$.
A higher distance value indicates more diverse features, while a lower value indicates feature redundancy.

To perform this analysis, we randomly select 1,000 images from different classes in the ImageNet-1K validation set and extract the high-dimensional features from each layer of different models, including the \convmae models, the ConvNeXt supervised model~\cite{liu2022convnet} and the MAE pre-trained ViT model~\cite{he2022masked}. We then compute the distance per layer for each image and average the values across all images. The results are plotted in Figure \ref{fig:feat_cos_dist}. The \convmae pre-trained ConvNeXt model exhibits a clear tendency towards feature collapse, consistent with our observations from the previous activation visualizations. This motivates us to consider ways to diversify the features during learning and prevent feature collapse.

\begin{table*}[t]
\vspace{-.2em}
\centering

\subfloat[
\textbf{Global aggregation $G(\cdot)$}. L2 Norm-based aggregation function produces the best result.
\label{tab:grn_spool}
]{
\begin{minipage}{0.3\linewidth}{\begin{center}
\tablestyle{1pt}{1.05}
\begin{tabular}{y{32}x{24}}
case & ft \\
\shline
g.avg.       & 83.7  \\
L1         & 84.3 \\
L2         & \baseline{\textbf{84.6}} \\
\end{tabular}
\end{center}}\end{minipage}
}
\hspace{0.5em}
\subfloat[
\textbf{Normalization operator, $N(\cdot)$}. Divisive normalization is an effective channel importance calibrator.
\label{tab:grn_cnorm}
]{
\centering
\begin{minipage}{0.35\linewidth}{\begin{center}
\tablestyle{4pt}{1.05}
\begin{tabular}{y{80}x{32}}
case & ft \\
\shline 
$(||X_{i}||-\mu)/\sigma$              & 84.5  \\
$1/\sum ||X_{i}||$                 & 83.8   \\
$||X_{i}||/\sum ||X_{i}||$              & \baseline{\textbf{84.6}} \\
\end{tabular}
\end{center}}\end{minipage}
}
\hspace{0.5em}
\subfloat[
\textbf{Residual connection} helps with GRN optimization and leads to better performance.
\label{tab:grn_residual}
]{
\begin{minipage}{0.29\linewidth}{\begin{center}
\tablestyle{1pt}{1.05}
\begin{tabular}{y{32}x{24}}
case & ft \\
\shline
{{w/o} skip}    & 84.0 \\
{{w/} skip}     & \baseline{\textbf{84.6}} \\
\multicolumn{2}{c}{~}\\
\end{tabular}
\end{center}}\end{minipage}
}
\\
\centering
\vspace{.1em}
\subfloat[
\textbf{Feature normalization}. GRN outperforms other normalizations through global contrasting.
\label{tab:grn_vs_norms}
]{
\centering
\begin{minipage}{0.3\linewidth}{\begin{center}
\tablestyle{1pt}{1.05}
\begin{tabular}{y{32}x{24}}
case & ft \\
\shline
\gc{Baseline}                       & \gc{83.7}   \\
LRN~\cite{krizhevsky2017imagenet}   & 83.2 \\
BN~\cite{ioffe2015batch}            & 80.5   \\
LN~\cite{ba2016layer}               & 83.8   \\ 
GRN                                 & \baseline{\textbf{84.6}} \\ 
\end{tabular}
\end{center}}\end{minipage}
}
\hspace{0.5em}
\subfloat[
\textbf{Feature re-weighting}. GRN does effective and efficient feature re-weighting without parameter overhead.
\label{tab:grn_vs_gates}
]{
\begin{minipage}{0.35\linewidth}{\begin{center}
\tablestyle{4pt}{1.05}
\begin{tabular}{y{48}x{32}x{32}}
case    & ft & \#param\\
\shline
\gc{Baseline}               &\gc{83.7} &\gc{89M}\\
SE~\cite{hu2018squeeze}     &84.4     &109M\\
CBAM~\cite{woo2018cbam}     &84.5     &109M\\
GRN                         &\baseline{\textbf{84.6}}     &\baseline{\textbf{89M}}\\
\multicolumn{2}{c}{~}\\
\end{tabular}
\end{center}}\end{minipage}
} 
\hspace{0.5em}
\subfloat[
\textbf{GRN in pre-training/fine-tuning}. To be effective, GRN should be used in both stages.
\label{tab:grn_ptft}
]{
\begin{minipage}{0.29\linewidth}{\begin{center}
\tablestyle{1pt}{1.05}
\begin{tabular}{y{32}x{24}}
case & ft \\
\shline
\gc{Baseline}     & \gc{83.7}     \\
{drop at ft.}        & 78.8          \\
{add at ft.}        & 80.6          \\
{both}            & \baseline{\textbf{84.6}} \\
\multicolumn{2}{c}{~}\\
\end{tabular}
\end{center}}\end{minipage}
}
\caption{\textbf{GRN ablations} with ConvNeXt-Base. We report fine-tuning accuracy on ImageNet-1K. Our final proposal is marked in \colorbox{baselinecolor}{gray}.}
\label{tab:ablations} 
\end{table*}

\paragraph{Approach.}
There are many mechanisms in the brain that promote neuron diversity. For example, lateral inhibition~\cite{hartline1956inhibition,campbell1968application} can help to sharpen the response of the activated neuron and increase the contrast and selectivity of individual neurons to the stimulus while also increasing the diversity of responses across the population of neurons. In deep learning,  this form of lateral inhibition can be implemented by response normalization~\cite{krizhevsky2017imagenet}. In this work, we introduce a new response normalization layer called global response normalization (GRN), which aims to increase the contrast and selectivity of channels. Given an input feature, $X \in R^{H\times W\times C}$, the proposed GRN unit consists of three steps: 1) global feature aggregation, 2) feature normalization, and 3) feature calibration.

First, we aggregate a spatial feature map $X_i$ into a vector $gx$ with a global function $\mathcal{G}(\cdot)$:
\begin{equation}
\mathcal{G}(X):= X \in \mathcal{R}^{H\times W\times C} \rightarrow gx \in \mathcal{R}^{C}.
\end{equation}
This can be viewed as a simple pooling layer. We experimented with different functions in Table~\ref{tab:grn_spool}.
Interestingly, global average pooling, a widely used feature aggregator~\cite{hu2018squeeze,woo2018cbam}, did not perform well in our case. 
Instead, we found that using norm-based feature aggregation, specifically, using L2-norm, resulted in better performance. This gives us a set of aggregated values $\mathcal{G}(X)=gx=\{||X_{1}||, ||X_{2}||, \ldots, ||X_{C}||\} \in \mathcal{R}^{C}$ where $\mathcal{G}(X)_i=||X_{i}||$ is a scalar that aggregates the statistics of the i-th channel.

Next, we apply a response normalization function $\mathcal{N}(\cdot)$ to the aggregated values. Concretely, we use a standard divisive normalization as follows,
\begin{equation}
\mathcal{N}(||X_{i}||):= ||X_{i}|| \in \mathcal{R} \rightarrow \frac{||X_{i}||}{\sum_{j=1,\ldots,C} ||X_{j}||}\in \mathcal{R},
\label{eq:grn_norm}
\end{equation}
where $||X_{i}||$ is the L2-norm of the $i$-th channel.~\footnote{To account for the increased number of channels at deeper layers,  in practice, we also scale the normalized value by the channel count $C$.} Intuitively, for the i-th channel, Eqn.~\ref{eq:grn_norm} computes its \emph{relative importance} compared to all the other channels. Similar to other forms of normalization~\cite{jarrett2009best,krizhevsky2017imagenet,vaswani2017attention}, this step creates a feature competition across channels by mutual inhibition. In Table~\ref{tab:grn_cnorm}, we also examine the use of other normalization functions and find that the simple divisive normalization works best, though standardization $(||X_{i}||-\mu)/\sigma$ yields similar results when applied to the same L2-norm aggregated values.

\begin{algorithm}[t]
\caption{Pseudocode of GRN in a PyTorch-like style.}
\label{alg:code}
\definecolor{codeblue}{rgb}{0.25,0.5,0.5}
\lstset{
  backgroundcolor=\color{white},
  basicstyle=\fontsize{7.2pt}{7.2pt}\ttfamily\selectfont,
  columns=fullflexible,
  breaklines=true,
  captionpos=b,
  commentstyle=\fontsize{7.2pt}{7.2pt}\color{codeblue},
  keywordstyle=\fontsize{7.2pt}{7.2pt},
  texcl=<true|false>
}
\begin{lstlisting}[language=python]
# gamma, beta: learnable affine transform parameters
# X: input of shape (N,H,W,C)

gx = torch.norm(X, p=2, dim=(1,2), keepdim=True)
nx = gx / (gx.mean(dim=-1, keepdim=True)+1e-6)
return gamma * (X * nx) + beta + X
\end{lstlisting}
\end{algorithm}

Finally, we calibrate the original input responses using the computed feature normalization scores:
\begin{equation}
X_{i} = X_{i} * \mathcal{N}(\mathcal{G}(X)_{i}) \in \mathcal{R}^{H\times W}
\label{eq:grn_equation}
\end{equation}
The core GRN unit is very easy to implement, requiring only three lines of code, and has no learnable parameters. The pseudo-code for the GRN unit is in Algorithm~\ref{alg:code}.

To ease optimization, we add two additional learnable parameters, $\gamma$ and $\beta$, and initialize them to zero. We also add a residual connection between the input and output of the GRN layer. The resulting final GRN block is $X_{i} = \gamma * X_{i} *  \mathcal{N}(\mathcal{G}(X)_{i}) + \beta + X_{i}$. This setup allows a GRN layer to initially perform an identity function and gradually adapt during training. The importance of residual connection is demonstrated in Table~\ref{tab:grn_residual}.

\begin{figure}[t]
\centering
\vstretch{0.95}{\includegraphics[width=0.45\textwidth]{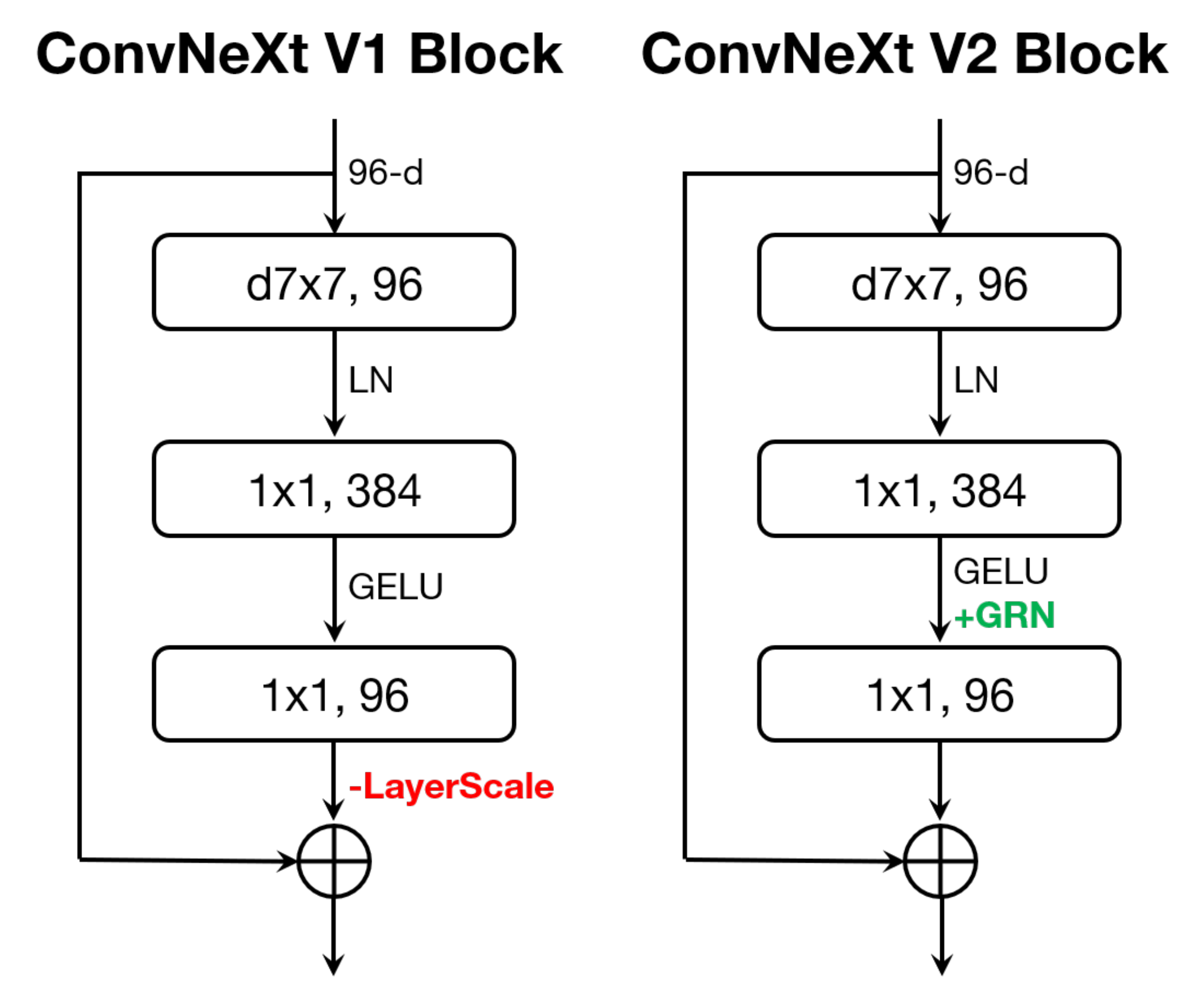}}
\vspace{-1em}
\caption{
\textbf{ConvNeXt Block Designs}. In ConvNeXt V2, we add the GRN layer after the dimension-expansion MLP layer and drop LayerScale~\cite{touvron2021going} as it becomes redundant.
}
\label{fig:block_design}
\vspace{-.5em}
\end{figure}

\paragraph{ConvNeXt V2.}
We incorporate the GRN layer into the original ConvNeXt block, as illustrated in Figure~\ref{fig:block_design}. We empirically found that LayerScale~\cite{touvron2021going} becomes unnecessary when GRN is applied and can be removed. Using this new block design, we create various models with varying efficiency and capacity, which we refer to as the ConvNeXt V2 model family. These models range from lightweight (\eg Atto~\cite{rw2019timm}) to compute-intensive (\eg Huge) ones. Detailed model configurations can be found in the appendix.

\paragraph{Impact of GRN.}
We now pre-train ConvNeXt V2 using the \convmae framework and evaluate the impact of GRN. From visualization in Figure~\ref{fig:feat_act} and cosine distance analysis in Figure~\ref{fig:feat_cos_dist}, we can observe that ConvNeXt V2 effectively mitigates the feature collapse issue. The cosine distance values are consistently high, indicating that feature diversity is maintained across layers.
This behavior is similar to that of the MAE pre-trained ViT model~\cite{he2022masked}.
Overall, this suggests that ConvNeXt V2 learning behavior can resemble ViT, under a similar masked image pre-training framework.

Next, we evaluate the fine-tuning performance.

\begin{center}\vspace{-.2em}
\tablestyle{4pt}{1.05}
\begin{tabular}{x{60}x{50}x{70}}
V1 + Sup, 300ep. & V1 + \convmae & V2 + \convmae \\
\shline
83.8 & 83.7 & 84.6
\end{tabular}\vspace{-.2em}
\end{center}

When equipped with GRN, the \convmae pre-trained model can significantly outperform the 300 epoch supervised counterpart. GRN improves the representation quality by enhancing the feature diversity, which was absent in the V1 model but has proven crucial for masked-based pre-training. Note this improvement is achieved \emph{without} adding additional parameter overhead or increased FLOPS.\footnote{The additional affine parameters $\gamma$/$\beta$ are negligible.}

\paragraph{Relation to feature normalization methods.}
Can other normalization layers~\cite{krizhevsky2017imagenet,ioffe2015batch,Ulyanov2016,ba2016layer,Wu2018} perform as well as the global response normalization (GRN) layer? 
In Table~\ref{tab:grn_vs_norms}, we compare GRN with the three widely used normalization layers: 
Local Response Normalization (LRN)~\cite{krizhevsky2017imagenet}, Batch Normalization (BN)~\cite{ioffe2015batch}, and Layer Normalization (LN)~\cite{ba2016layer}.
We observe that only GRN can significantly outperform the supervised baseline. LRN lacks global context as it only contrasts channels within nearby neighbors. BN normalizes spatially along the batch axis, which is unsuitable for masked inputs. LN implicitly encourages feature competition through global mean and variance standardization but does not work as well as GRN.

\paragraph{Relation to feature gating methods.}
Another way to enhance competition across neurons is to use dynamic feature gating methods~\cite{hu2018squeeze,park2018bam,woo2018cbam,Wang_2020_CVPR,yang2020gated}.
In Table~\ref{tab:grn_vs_gates}, we compare our GRN with two classic gating layers: squeeze-and-excite (SE)~\cite{hu2018squeeze} and convolutional block attention module (CBAM)~\cite{woo2018cbam}. SE focuses on channel gating, while CBAM focuses on spatial gating. Both modules can increase the contrast of individual channels, similar to what GRN does. GRN is much simpler and more efficient as it does not require additional parameter layers (such as MLPs).

\paragraph{The role of GRN in pre-training/fine-tuning.}
Finally, we examine the importance of GRN in pre-training and fine-tuning. We present results in Table~\ref{tab:grn_ptft} where we either remove GRN from fine-tuning or add newly initialized GRN only at the time of fine-tuning. Either way, we observe a significant performance degradation, suggesting that keeping GRN in both pre-training and fine-tuning is important.

\definecolor{Highlight}{HTML}{39b54a}  %
\renewcommand{\hl}[1]{\textcolor{Highlight}{#1}}

\begin{table}
\centering
\vspace{-.5em}
\tablestyle{3pt}{1.2}
\begin{tabular}{l y{42}x{32}x{32}y{42}}
\multirow{1}{*}{Backbone} 
& \multicolumn{1}{l}{Method} & \multicolumn{1}{c}{\#param} & \multicolumn{1}{c}{FLOPs} & \multicolumn{1}{c}{Val acc.} \\
\shline
ConvNeXt V1-B      & Supervised & 89M         &15.4G    &83.8 \\
ConvNeXt V1-B      & \convmae   & 89M         &15.4G    &83.7 \\
ConvNeXt V2-B      & Supervised & 89M         &15.4G    &84.3 ($+$0.5) \\
\gr
ConvNeXt V2-B      & \convmae      & 89M         &15.4G    &\textbf{84.6} \textcolor{Highlight}{($+$\textbf{0.8})}\\
\hline
ConvNeXt V1-L      & Supervised & 198M        &34.4G    &84.3\\
ConvNeXt V1-L      & \convmae   & 198M        &34.4G    &84.4\\
ConvNeXt V2-L      & Supervised & 198M        &34.4G    &84.5 ($+$0.2)\\
\gr
ConvNeXt V2-L      & \convmae      & 198M        &34.4G    &\textbf{85.6} \textcolor{Highlight}{($+$\textbf{1.3})}\\
\end{tabular}
\caption{\textbf{Co-design matters}. 
When the architecture and the learning framework are co-designed and used together, masked image pre-training becomes effective for ConvNeXt. We report the fine-tuning performance from 800 epoch \convmae pre-trained models. The relative improvement is bigger with a larger model.
}
\vspace{-1em}
\label{tab:codesign_exp}
\end{table}

\section{ImageNet Experiments}

In this section, we present and analyze two key proposals, the \convmae \emph{pre-training framework} and ConvNeXt V2 \emph{architecture}, which are co-designed to make masked-based self-supervised pre-training successful.
We show these designs synergize well and provide a strong foundation for scaling the model to various sizes. Additionally, we compare our approach to previous masked image modeling approaches through experiments. Furthermore, we show that our largest ConvNeXt V2 Huge model, which has been pre-trained using the \convmae framework and fine-tuned on the ImageNet-22K dataset, can achieve a new state-of-the-art of 88.9\% top-1 accuracy on the ImageNet-1K dataset, using only publicly available data.

\begin{table}
\vspace{-.5em}
\tablestyle{3pt}{1.2}
\begin{tabular}{l y{48}x{32}x{32}x{36}}
\multirow{1}{*}{Backbone} 
& \multicolumn{1}{l}{Method} & \multicolumn{1}{c}{\#param} & \multicolumn{1}{c}{PT epoch} & \multicolumn{1}{c}{FT acc.} \\
\shline
ViT-B        & BEiT         & 88M    & 800       & 83.2 \\ %
ViT-B         & MAE        & 88M    & 1600      & 83.6 \\ %
Swin-B      & SimMIM        & 88M    & 800       & 84.0 \\ %
\gr
ConvNeXt V2-B       & \convmae   & 89M    & 800       & 84.6 \\ %
\gr
ConvNeXt V2-B       & \convmae   & 89M    & 1600      & \underline{\textbf{84.9}} \\
\hline
ViT-L        & BEiT        & 307M   & 800       & 85.2 \\ %
ViT-L        & MAE         & 307M   & 1600      & \underline{85.9} \\ %
Swin-L       & SimMIM        & 197M   & 800       & 85.4 \\ %
\gr
ConvNeXt V2-L      & \convmae & 198M   & 800         & 85.6 \\ %
\gr
ConvNeXt V2-L      & \convmae & 198M   & 1600        & \textbf{85.8} \\
\hline
ViT-H         & MAE         & 632M    & 1600        & \underline{86.9} \\
Swin V2-H      & SimMIM     & 658M    & 800      & 85.7 \\
\gr
ConvNeXt V2-H       & \convmae & 659M    & 800        & 85.8 \\
\gr
ConvNeXt V2-H       & \convmae & 659M    & 1600       & \textbf{86.3} \\
\end{tabular}
\caption{\textbf{Comparisons with previous masked image modeling approaches}. 
The pre-training data is the IN-1K training set. All self-supervised methods are benchmarked by the end-to-end fine-tuning performance with an image size of 224. We underline the highest accuracy for each model size and bold our best results.
}
\vspace{-1em}
\label{tab:imagenet_e2e}
\end{table}

\paragraph{Co-design matters.}
In this paper, we conduct a unique study that involves \emph{co-designing} both the self-supervised learning framework (\convmae) and the model architecture improvement (GRN layer), through an empirical study of their learning behavior. The results presented in Table~\ref{tab:codesign_exp} demonstrate the importance of this approach. 

We found that using the \convmae framework without modifying the model architecture has a limited impact on representation learning quality. Similarly, the new GRN layer has a rather small effect on performance under the supervised setup. However, the combination of the two results in a significant improvement in fine-tuning performance. This supports the idea that both the model and learning framework should be considered together, particularly when it comes to self-supervised learning.

\paragraph{Model scaling.}
In this study, we evaluated a range of 8 models with different sizes, from a low-capacity 3.7M Atto model to a high-capacity 650M Huge model. We pre-trained these models using the proposed \convmae framework and compared the fine-tuning results to the fully supervised counterparts. 

The results, shown in Figure~\ref{fig:teaser}, demonstrate strong model scaling behavior, with consistently improved performance over the supervised baseline across all model sizes. This is the first time the benefit of masked image modeling has been demonstrated in such a broad model spectrum, both in terms of effectiveness and efficiency. The complete tabulated results can be found in the appendix.

\paragraph{Comparisons with previous methods.}
We compare our approach to previous masked auto-encoder methods~\cite{Bao2021,he2022masked,xie2022simmim}, which were all designed for transformer-based models. The results are summarized in Table~\ref{tab:imagenet_e2e}. Our framework outperforms the Swin transformer pre-trained with SimMIM~\cite{xie2022simmim} \emph{across all model sizes}. Compared to the plain ViT pre-trained with MAE~\cite{he2022masked}, our approach performs similarly up to the Large model regime, despite using much fewer parameters (198M vs 307M). However, in the huge model regime, our approach slightly lagged behind. This might be because a huge ViT model can benefit more from self-supervised pre-training. As we will see next, the gap might be closed with additional intermediate fine-tuning.

\definecolor{Highlight}{HTML}{39b54a}  %
\renewcommand{\hl}[1]{\textcolor{Highlight}{#1}}

\begin{table}
\centering
\vspace{-.5em}
\tablestyle{5pt}{1.2}
\begin{tabular}{l y{60}x{24}x{24}x{24}x{24}}
\multirow{1}{*}{Type}
&
\multirow{1}{*}{Backbone} 
& \multicolumn{1}{c}{size} & \multicolumn{1}{c}{\#param} & \multicolumn{1}{c}{FLOPS} & \multicolumn{1}{c}{Val acc.} \\
\shline
\multirow{2}{*}{Conv}   &Efficient V2-XL & 480\textsuperscript{2}    & 208M       &94.0G       &87.3 \\
                        &ConvNeXt V1-XL     & 384\textsuperscript{2}    & 350M       &179.0G      &87.8 \\
\hline
\multirow{3}{*}{Hybrid} &CoAtNet-4          & 512\textsuperscript{2}    & 275M       &360.9G    &88.1 \\
                        &MaxViT-XL          & 384\textsuperscript{2}    & 475M       &293.7G    &88.5 \\
                        &MaxViT-XL          & 512\textsuperscript{2}    & 475M       &535.2G    &88.7 \\
\hline
\multirow{2}{*}{Trans}  &MViTV2-H           & 384\textsuperscript{2}    & 667M       &388.5G    &88.6 \\
                        &MViTV2-H           & 512\textsuperscript{2}    & 667M       &763.5G    &88.8 \\
\hline
\gr
                       &ConvNeXt V2-H      & 384$^{2}$    & 659M        & 337.9G        &88.7\\
\gr
\multirow{-2}{*}{Conv} &ConvNeXt V2-H      & 512$^{2}$    & 659M        & 600.7G        &\textbf{88.9}\\
\end{tabular}
\caption{\textbf{ImageNet-1K fine-tuning results using IN-21K labels}. The ConvNeXt V2 Huge model equipped with the \convmae pre-training outperforms other architectures and sets a new state-of-the-art accuracy of 88.9\% among methods using public data only.
}
\vspace{-1.5em}
\label{tab:imagenet_sota}
\end{table}

\paragraph{ImageNet-22K intermediate fine-tuning.}
We also present ImageNet-22K intermediate fine-tuning results~\cite{Bao2021}. The training process involves three steps: 1) \convmae pre-training, 2) ImageNet-22K fine-tuning, and 3) ImageNet-1K fine-tuning. We use $384^{2}$ resolution images for pre-training and fine-tuning~\cite{hu2022exploring}. We compare our results to the state-of-the-art architecture designs, including convolution-based~\cite{tan2021efficientnetv2,liu2022convnet}, transformer-based~\cite{fan2021multiscale}, and hybrid designs~\cite{dai2021coatnet,tu2022maxvit}. All these results were trained with ImageNet-22K supervised labels. The results are summarized in Table~\ref{tab:imagenet_sota}. Our method, using a convolution-based architecture, sets a new state-of-the-art accuracy using publicly available data only (\ie ImageNet-1K and ImageNet-22K).

\section{Transfer Learning Experiments}
We now benchmark the transfer learning performance. First, we evaluate the impact of our co-design, \ie comparing ConvNeXt V1~+~supervised \textit{vs.} ConvNeXt V2~+~FCMAE.
We also directly compare our approach with Swin transformer models pre-trained with SimMIM~\cite{xie2022simmim}. 
The training and testing details are provided in the appendix.

\paragraph{Object detection and segmentation on COCO.}
We fine-tune Mask R-CNN \cite{He2017} on the COCO dataset \cite{Lin2014} and report the detection mAP\textsuperscript{box} and the segmentation mAP\textsuperscript{mask} on the COCO val2017 set. The results are shown in Table~\ref{tab:coco_transfer}. We see a gradual improvement as our proposals are applied. From V1 to V2, the GRN layer is newly introduced and enhances performance. Upon this, the model further benefits from better initialization when moving from supervised to FCMAE-based self-supervised learning. The best performances are achieved when both are applied together. Additionally, our final proposal, ConvNeXt V2 pre-trained on FCMAE, outperforms the Swin transformer counterparts across all model sizes, with the largest gap achieved in the huge model regime.

\begin{table}
\begin{center}\vspace{-.1em}
\tablestyle{3pt}{1.3}
\scriptsize{
\begin{tabular}{l y{27}x{18}x{14}x{14}x{14}x{16}x{16}x{16}x{16}}
Backbone & Method & FLOPS & AP$^\textsuperscript{box}$ & AP$^{\textsuperscript{box}}_{50}$ & AP$^{\textsuperscript{box}}_{75}$ & AP$^\textsuperscript{mask}$ & AP$^{\textsuperscript{mask}}_{50}$ & AP$^{\textsuperscript{mask}}_{75}$ \\
\shline
ConvNeXt V1-B & Supervised & 486G & 50.3 & 71.6 & 56.1 & 44.9 & 68.5 & 48.8 \\
ConvNeXt V2-B & Supervised & 486G & 51.0 & 72.4 & 56.6 & 45.6 & 69.5 & 49.7 \\
Swin-B & SimMIM & 497G & 52.3 & $-$ & $-$ & $-$ & $-$ & $-$  \\
\gr
ConvNeXt V2-B & \convmae & 486G & \textbf{52.9} & 72.6 & 58.9 & \textbf{46.6} & 70.0 & 51.1 \\
\hline
ConvNeXt V1-L & Supervised & 875G & 50.6 & 71.5 & 56.3 & 45.1 & 68.7 & 49.2 \\
ConvNeXt V2-L & Supervised & 875G & 51.5 & 72.5 & 57.3 & 45.8 & 69.4 & 49.9 \\
Swin-L & SimMIM & 904G & 53.8 & $-$ & $-$ & $-$ & $-$ & $-$ \\
\gr
ConvNeXt V2-L & \convmae & 875G & \textbf{54.4} & 73.9 & 60.4 & \textbf{47.7} & 71.4 & 52.3 \\
\hline
Swin V2-H & SimMIM & $-$ & 54.4 & $-$ & $-$ & $-$ & $-$ & $-$ \\
\gr
ConvNeXt V2-H & \convmae & 2525G & \textbf{55.7} & 75.2 & 61.8 & \textbf{48.9} & 72.8 & 53.6
\end{tabular}\vspace{-.5em}
}
\end{center}
\caption{\label{tab:coco_transfer}\textbf{COCO object detection and instance segmentation results} using Mask-RCNN. FLOPS are calculated with image size (1280, 800). Swins’ results are from ~\cite{xie2022simmim}. All COCO fine-tuning experiments rely on ImageNet-1K pre-trained models.}
\end{table}

\begin{table}
\begin{center}\vspace{-.1em}
\tablestyle{4pt}{1.2}
\begin{tabular}{l y{52}x{15}x{19}x{23}x{20}}
Backbone & Method & input & mIoU & \#param & FLOPS\\
\shline
ConvNeXt V1-B & Supervised & 512\textsuperscript{2} & 49.9 & 122M & 1170G \\
ConvNeXt V2-B & Supervised & 512\textsuperscript{2} & 50.5 & 122M & 1170G \\
Swin-B & SimMIM & 512\textsuperscript{2} & \textbf{52.8} & 121M & 1181G \\
\gr
ConvNeXt V2-B & \convmae & 512\textsuperscript{2} & 52.1 & 122M & 1170G   \\
\hline
ConvNeXt V1-L & Supervised & 512\textsuperscript{2} & 50.5 & 235M & 1573G \\
ConvNeXt V2-L & Supervised & 512\textsuperscript{2} & 51.6 & 235M & 1573G  \\
Swin-L & SimMIM & 512\textsuperscript{2} & 53.5 & 234M  & 1601G \\
\gr
ConvNeXt V2-L & \convmae & 512\textsuperscript{2} & \textbf{53.7} & 235M & 1573G  \\
\hline
Swin V2-H & SimMIM & 512\textsuperscript{2} & 54.2 & $-$ & $-$  \\
\gr
ConvNeXt V2-H & \convmae & 512\textsuperscript{2} & \textbf{55.0} & 707M & 3272G \\
\hline
\gr
ConvNeXt V2-H & \convmae, \emph{22K ft} & 640\textsuperscript{2} & \textbf{57.0} & 707M & 5113G
\end{tabular}\vspace{-1em}
\end{center}
\caption{\label{tab:ade_transfer}\textbf{ADE20K semantic segmentation results} using UPerNet. Swins’ results are from ~\cite{xie2022simmim}. FLOPS are based on input sizes of (2048, 512) or (2560, 640). All ADE20K fine-tuning experiments rely on ImageNet-1K pre-trained model except \convmae, \textit{22K ft}, in which case the ImageNet-1K pre-training is followed by ImageNet-22K supervised fine-tuning.}
\end{table}

\paragraph{Semantic segmentation on ADE20K.}
To summarize, we conduct experiments on the ADE20K \cite{Zhou2019} semantic segmentation task using the UperNet framework\cite{Xiao2018}. Our results show a similar trend to the object detection experiments, and our final model significantly improves over the V1 supervised counterparts. It also performs on par with the Swin transformer in the base and large model regimes but outperforms Swin in the huge model regime.

\section{Conclusion} In this paper, we introduce a new ConvNet model family called ConvNeXt V2 that covers a broader range of complexity. While the architecture has minimal changes, it is specifically designed to be more suitable for self-supervised learning. Using our fully convolutional masked autoencoder pre-training, we can significantly improve the performance of pure ConvNets across various downstream tasks, including ImageNet classification, COCO object detection, and ADE20K segmentation.

\paragraph{Acknowledgments.} We thank Ross Wightman for the initial design of the small-compute ConvNeXt model variants and the associated training recipe. We also appreciate the helpful discussions and feedback provided by Kaiming He.

\clearpage

\appendix
\section*{\Large{Appendix}}
This appendix provides implementation details, including model configurations, pre-training and fine-tuning recipes, and sparse and dense encoding methods for FCMAE pre-training (see \S \ref{sec:impl}). In \S \ref{sec:comp}, we present complete fine-tuning accuracy comparisons between ConvNeXt V1 and V2 on ImageNet 1K and 22K. In \S \ref{sec:anal}, we perform analyses on the efficiency of sparse encoding and general feature analysis using the class selectivity index. Finally, in \S \ref{sec:exps}, we conduct additional ablation studies on the masking ratio and GRN component analysis. We also compare FCMAE (masked image modeling) with MoCo V3 (contrastive learning).

\section{Implementation Details}
\label{sec:impl}
\subsection{ConvNeXt V2 model configurations}
The basic models, \ie, Tiny (28M), Base (89M) and Large (198M), follow the same configurations of the stage, block (B), and channel (C) settings of the ConvNeXt V1~\cite{liu2022convnet}.

\begin{itemize}\small
    \item ConvNeXt V2-T: $C$=96, $B$=(3, 3, 9, 3)
    \item ConvNeXt V2-B: $C$=128, $B$=(3, 3, 27, 3)
    \item ConvNeXt V2-L: $C$=192, $B$=(3, 3, 27, 3)
\end{itemize}

Given the same definitions above, we scale the model to provide a broad model size spectrum, targeting versatile scenarios.
First, to obtain efficient models, we scale down as follows:

\begin{itemize}\small
    \item ConvNeXt V2-A: $C$=40, $B$=(2, 2, 6, 2)
    \item ConvNeXt V2-F: $C$=48, $B$=(2, 2, 6, 2)
    \item ConvNeXt V2-P: $C$=64, $B$=(2, 2, 6, 2)
    \item ConvNeXt V2-N: $C$=80, $B$=(2, 2, 8, 2)
\end{itemize}

A, F, P, N denote Atto (3.7M), Femto (5.2M), Pico (9.1M), and Nano (15.6M) models designed originally in~\cite{rw2019timm}. Next, to introduce the large-capacity variant, we scale up as follows:

\begin{itemize}\small
    \item ConvNeXt V2-H: $C$=352, $B$=(3, 3, 27, 3)
\end{itemize}

H denotes Huge (659M) model, which is newly presented in this work.

\subsection{ImageNet Experiments}

\paragraph{Pre-training}
All models share the same pre-training setup, as noted in Table~\ref{tab:impl_mae_pretrain}. 
We use the linear \textit{lr} scaling rule \cite{Goyal2017}: \textit{lr} = \textit{base\_lr}$\times$batchsize / 256.

\paragraph{ImageNet-1K fine-tuning}
As the learning capacity varies by model size, we adopt different fine-tuning recipes for each model.
We summarize them in Table~\ref{tab:impl_mae_finetune_atto}, \ref{tab:impl_mae_finetune_tiny} and~\ref{tab:impl_mae_finetune}.
We see longer fine-tuning epochs help small models.
We adopt two different learning-rate layer decay strategies in this work: group-wise~\cite{liu2022convnet}, where we treat three sequential layers as a single ``layer'' and use the same decaying value for them, and the layer-wise~\cite{Bao2021}, where we assign a distinct value for each layer, both following the standard decaying rule. The default is a layer-wise strategy, but we apply the group-wise decaying strategy to Base and Large models.

\paragraph{ImageNet-22K intermediate fine-tuning}
We conduct ImageNet-22K intermediate fine-tuning with the FCMAE-pretrained ConvNeXt models.
We use nano, tiny, base, large, and huge models.
The setups are summarized in Table~\ref{tab:impl_mae_finetune_im22k} and \ref{tab:impl_mae_finetune_im22k_im1k}.
Similarly, using larger layer-wise learning rate decay values for small models is helpful.

\paragraph{Sparse encoding implementations.}
We propose two possible implementations to enable FCMAE pre-training: 1) sparse encoding using sparse convolution~\cite{graham2017submanifold,graham20183d,choy20194d} supported by external libraries~\cite{choy20194d,spconv2022}, and 2) simulating sparse encoding with the masked dense convolution, which can be easily implemented by applying binary masks \emph{before and after} the standard convolution operation.
As they produce numerically identical outputs, both can be adopted depending on different use cases. 
In this work, we adopt sparse encoding on the GPU environment, where we use MinkowskiEngine library~\cite{choy20194d} and PyTorch framework~\cite{paszke2017automatic}; we use dense masked conv based encoding on TPU accelerators using Jax~\cite{jax2018github}. 
The experiments in the main paper are all conducted on TPU (v3-256) pods and we release a PyTorch reproduction.

\subsection{Object detection and segmentation on COCO}

For COCO experiments, we use the MMDetection~\cite{mmdetection} toolbox and the final model weights from ImageNet-1K pre-training as network initializations. All models are trained with a 3x schedule (36 epochs) and a batch size of 32. We utilize an AdamW optimizer \cite{Loshchilov2019} with a learning rate of 1e-4, a weight decay of 0.05 and sweep layer-wise learning rate decay in \{0.9, 0.95\}, stochastic depth rate in \{0.2, 0.3, 0.4, 0.5\}. We employ a large-scale jittering augmentation \cite{Ghiasi2021} (1024$\times$1024 resolution, scale range [0.1, 2.0]). We use single-scale testing with soft-NMS \cite{bodla2017soft} during inference.

\begin{table}[t]
\tablestyle{6pt}{1.02}
\scriptsize
\begin{tabular}{y{84}|y{80}}
config & value \\
\shline
optimizer & AdamW \cite{Loshchilov2019} \\
base learning rate & 1.5e-4 \\
weight decay & 0.05 \\
optimizer momentum & $\beta_1, \beta_2{=}0.9, 0.95$ \cite{Chen2020c} \\
batch size & 4096 \\
learning rate schedule & cosine decay \cite{Loshchilov2016} \\
warmup epochs \cite{Goyal2017} & 40 \\
training epochs & 800 or 1600 \\
augmentation & RandomResizedCrop \\
\end{tabular}
\vspace{-1.em}
\caption{\textbf{Pre-training setting.}}
\label{tab:impl_mae_pretrain}
\end{table}

\begin{table}[t]
\tablestyle{6pt}{1.02}
\scriptsize
\begin{tabular}{y{84}|y{80}}
config & value \\
\shline
optimizer & AdamW \\
base learning rate & 2e-4 \\
weight decay & 0.05 (F), 0.3 (A/P/N) \\
optimizer momentum & $\beta_1, \beta_2{=}0.9, 0.999$ \\
layer-wise lr decay \cite{Clark2020,Bao2021} & 0.9 \\
batch size & 1024 \\
learning rate schedule & cosine decay \\
warmup epochs & 0 \\
training epochs & 600 \\
augmentation & RandAug (9, 0.5) \cite{Cubuk2020} \\
label smoothing \cite{Szegedy2016a} & 0.2 \\
mixup \cite{Zhang2018a} & 0.0 (A), 0.3 (F/P), 0.5 (N)\\
cutmix \cite{Yun2019} & 0.0 (A), 0.3 (F/P), 0.5 (N) \\
drop path \cite{Huang2016deep} & 0.1 (A/N), 0.0 (F/P),  \\
head init \cite{liu2022convnet} & 0.001 \\
ema  & 0.9999 \\
\end{tabular}
\vspace{-1.em}
\caption{\textbf{End-to-end IN-1K fine-tuning setting for Atto (A), Femto (F), Pico (P) and Nano (N) models.}
}
\label{tab:impl_mae_finetune_atto} 
\end{table}

\begin{table}[t]
\tablestyle{6pt}{1.02}
\scriptsize
\begin{tabular}{y{84}|y{80}}
config & value \\
\shline
optimizer & AdamW \\
base learning rate & 8e-4 \\
weight decay & 0.05 \\
optimizer momentum & $\beta_1, \beta_2{=}0.9, 0.999$ \\
layer-wise lr decay \cite{Clark2020,Bao2021} & 0.9 \\
batch size & 1024 \\
learning rate schedule & cosine decay \\
warmup epochs & 40 \\
training epochs & 300 \\
augmentation & RandAug (9, 0.5) \cite{Cubuk2020} \\
label smoothing \cite{Szegedy2016a} & 0.1 \\
mixup \cite{Zhang2018a} & 0.8 \\
cutmix \cite{Yun2019} & 1.0 \\
drop path \cite{Huang2016deep} & 0.2 \\
head init \cite{liu2022convnet} & 0.001 \\
ema  & 0.9999 \\
\end{tabular}
\vspace{-1.em}
\caption{\textbf{End-to-end IN-1K fine-tuning setting for Tiny model.
}}
\label{tab:impl_mae_finetune_tiny}
\end{table}

\begin{table}[t]
\tablestyle{6pt}{1.02}
\scriptsize
\begin{tabular}{y{84}|y{80}}
config & value \\
\shline
optimizer & AdamW \\
base learning rate & 6.25e-3 (B/L), 1.25e-3 (H) \\
weight decay & 0.05 \\
optimizer momentum & $\beta_1, \beta_2{=}0.9, 0.999$ \\
layer-wise lr decay \cite{Clark2020,Bao2021} & 0.6 (B/L), 0.75 (H) \\
batch size & 1024 \\
learning rate schedule & cosine decay \\
warmup epochs & 20 (B/L), 10 (H) \\
training epochs & 100 (B/L), 50 (H) \\
augmentation & RandAug (9, 0.5) \cite{Cubuk2020} \\
label smoothing \cite{Szegedy2016a} & 0.1 \\
mixup \cite{Zhang2018a} & 0.8 \\
cutmix \cite{Yun2019} & 1.0 \\
drop path \cite{Huang2016deep} & 0.1 (B), 0.2 (L), 0.3 (H) \\
head init \cite{liu2022convnet} & 0.001 \\
ema  & 0.9999 \\
\end{tabular}
\vspace{-1.em}
\caption{\textbf{End-to-end IN-1K fine-tuning setting for Base (B), Large (L), and Huge (H) models.}}
\label{tab:impl_mae_finetune} 
\vspace{-.5em}
\end{table}

\begin{table}[t]
\tablestyle{6pt}{1.02}
\scriptsize
\begin{tabular}{y{84}|y{96}}
config & value \\
\shline
optimizer & AdamW \\
base learning rate & 2.5e-4 \\
weight decay & 0.05 \\
optimizer momentum & $\beta_1, \beta_2{=}0.9, 0.999$ \\
layer-wise lr decay \cite{Clark2020,Bao2021} & 0.8 (B/L/H), 0.9 (N/T) \\
batch size & 4096 \\
learning rate schedule & cosine decay \\
warmup epochs & 5 \\
training epochs & 90 \\
augmentation & RandAug (9, 0.5) \cite{Cubuk2020} \\
label smoothing \cite{Szegedy2016a} & 0.1 \\
mixup \cite{Zhang2018a} & 0.8 \\
cutmix \cite{Yun2019} & 1.0 \\
drop path \cite{Huang2016deep} & 0.(N/T), 0.1 (B/L), 0.3 (H) \\
head init \cite{liu2022convnet} & 0.001 \\
ema  & None \\
\end{tabular}
\vspace{-1.em}
\caption{\textbf{End-to-end IN-22K intermediate fine-tuning settings.}}
\label{tab:impl_mae_finetune_im22k} 
\end{table}

\begin{table}[t]
\tablestyle{6pt}{1.02}
\scriptsize
\begin{tabular}{y{84}|y{96}}
config & value \\
\shline
optimizer & AdamW \\
base learning rate & 2.5e-5 \\
weight decay & 1e-8 \\
optimizer momentum & $\beta_1, \beta_2{=}0.9, 0.999$ \\
layer-wise lr decay \cite{Clark2020,Bao2021} & 0.8 (B/L), 0.85 (H), 0.9 (N/T) \\
batch size & 512 \\
learning rate schedule & cosine decay \\
warmup epochs & None \\
training epochs & 30 (B/L/H), 90 (N/T) \\
augmentation & RandAug (9, 0.5) \cite{Cubuk2020} \\
label smoothing \cite{Szegedy2016a} & 0.1 \\
mixup \cite{Zhang2018a} & None \\
cutmix \cite{Yun2019} & None \\
drop path \cite{Huang2016deep} & 0.1(N/T), 0.2 (B), 0.3 (L), 0.5(H) \\
head init \cite{liu2022convnet} & 0.001 \\
ema  & 0.9999 (N/T/B/L), None (H) \\
\end{tabular}
\vspace{-1.em}
\caption{\textbf{End-to-end IN-1K fine-tuning settings (after IN-22K intermediate fine-tuning).}}
\label{tab:impl_mae_finetune_im22k_im1k} 
\vspace{-1.em}
\end{table}

\subsection{Semantic segmentation in ADE20K}

For ADE20K experiments, we use the MMSegmentation ~\cite{mmseg2020} toolbox. We use an AdamW optimizer \cite{Loshchilov2019} with the following hyperparameters: a weight decay of 0.05, a batch size of 16 and sweep layer-wise decay rate \{0.8, 0.9\}, learning rate \{1e-4, 2e-4, 3e-4\}, stochastic depth rate \{0.1, 0.2, 0.3, 0.4\}. All models are trained for 160K iterations with an input resolution of 512$\times$512. In inference, a multi-scale test using resolutions that are [0.75,0.875,1.0,1.125,1.25] of 512$\times$2048 is employed. 

Similar to~\cite{xie2022simmim}, we initialized the segmentation models using model weights after supervised fine-tuning on ImageNet-1K, as we found its performance superior to using the self-supervised pre-trained weights directly.

\definecolor{Highlight}{HTML}{39b54a}  %
\renewcommand{\hl}[1]{\textcolor{Highlight}{#1}}

\begin{table}
\centering
\vspace{-.5em}
\tablestyle{3pt}{1.2}
\begin{tabular}{l y{42}x{32}x{32}y{42}}
\multirow{1}{*}{Backbone} 
& \multicolumn{1}{l}{Method} & \multicolumn{1}{c}{\#param} & \multicolumn{1}{c}{FLOPs} & \multicolumn{1}{c}{Val acc.} \\
\shline
ConvNeXt V1-A      & Supervised     & 3.7M         &0.55G       &75.7 \\
ConvNeXt V2-A      & Supervised     & 3.7M         &0.55G       &76.2 ($+$0.5)\\
\gr
ConvNeXt V2-A      & \convmae       & 3.7M         &0.55G       &\textbf{76.7} \textcolor{Highlight}{($+$\textbf{1.0})}\\
\hline
ConvNeXt V1-F      & Supervised     & 5.2M         &0.78G       &77.5 \\
ConvNeXt V2-F      & Supervised     & 5.2M         &0.78G       &78.0 ($+$0.5)\\
\gr
ConvNeXt V2-F      & \convmae       & 5.2M         &0.78G       &\textbf{78.5} \textcolor{Highlight}{($+$\textbf{1.0})}\\
\hline
ConvNeXt V1-P      & Supervised     & 9.1M         &1.37G       &79.5 \\
ConvNeXt V2-P      & Supervised     & 9.1M         &1.37G       &79.7 ($+$0.2)\\
\gr
ConvNeXt V2-P      & \convmae       & 9.1M         &1.37G       &\textbf{80.3} \textcolor{Highlight}{($+$\textbf{0.8})}\\
\hline
ConvNeXt V1-N      & Supervised     & 15.6M        &2.45G       &80.8 \\
ConvNeXt V2-N      & Supervised     & 15.6M        &2.45G       &81.2 ($+$0.4)\\
\gr
ConvNeXt V2-N      & \convmae       & 15.6M        &2.45G       &\textbf{81.9} \textcolor{Highlight}{($+$\textbf{1.1})}\\
\hline
ConvNeXt V1-T      & Supervised     & 28.6M        &4.47G       &82.1 \\
ConvNeXt V2-T      & Supervised     & 28.6M        &4.47G       &82.5 ($+$0.4)\\
\gr
ConvNeXt V2-T      & \convmae       & 28.6M        &4.47G       &\textbf{83.0} \textcolor{Highlight}{($+$\textbf{0.9})}\\
\hline
ConvNeXt V1-B      & Supervised     & 89M          &15.4G       &83.8 \\
\gc{ConvNeXt V1-B} & \gc{\convmae}  &\gc{89M}      &\gc{15.4G}  &\gc{83.7} \\
ConvNeXt V2-B      & Supervised     & 89M          &15.4G       &84.3 ($+$0.5) \\
\gr
ConvNeXt V2-B      & \convmae       & 89M          &15.4G       &\textbf{84.9} \textcolor{Highlight}{($+$\textbf{1.1})}\\
\hline
ConvNeXt V1-L      & Supervised     & 198M         &34.4G       &84.3\\
\gc{ConvNeXt V1-L} & \gc{\convmae}  &\gc{198M}     &\gc{34.4G}  &\gc{84.4}\\
ConvNeXt V2-L      & Supervised     & 198M         &34.4G       &84.5 ($+$0.2)\\
\gr
ConvNeXt V2-L      & \convmae       & 198M         &34.4G       &\textbf{85.8} \textcolor{Highlight}{($+$\textbf{1.5})}\\
\hline
ConvNeXt V2-H      & \convmae       & 660M         &115G     &\textbf{86.3} \\
\end{tabular}
\caption{\textbf{ImageNet-1K fine-tuning results} with a single 224$\times$224 crop. The improvement over the V1 supervised model is shown in parentheses.}
\label{tab:v1_v2_comp_im1k}
\end{table}

\definecolor{Highlight}{HTML}{39b54a}  %
\renewcommand{\hl}[1]{\textcolor{Highlight}{#1}}

\begin{table}
\centering
\vspace{-.5em}
\tablestyle{3pt}{1.2}
\begin{tabular}{l x{42}x{32}x{32}y{42}}
\multirow{1}{*}{Backbone} 
& \multicolumn{1}{c}{image size} & \multicolumn{1}{c}{\#param} & \multicolumn{1}{c}{FLOPs} & \multicolumn{1}{c}{Val acc.} \\
\shline
\gr
ConvNeXt V2-N      & $224^2$     & 15.6M         &2.45G       &\textbf{82.1} \\
\gr
ConvNeXt V2-N      & $384^2$     & 15.6M         &7.21G       &\textbf{83.4} \\
\hline
ConvNeXt V1-T      & $224^2$     & 28.6M         &4.47G       &82.9 \\
\gr
ConvNeXt V2-T      & $224^2$     & 28.6M         &4.47G       &\textbf{83.9}\textcolor{Highlight}{($+$\textbf{1.0})} \\
ConvNeXt V1-T      & $384^2$     & 28.6M         &13.1G       &84.1 \\
\gr
ConvNeXt V2-T      & $384^2$     & 28.6M         &13.1G       &\textbf{85.1}\textcolor{Highlight}{($+$\textbf{1.0})} \\
\hline
ConvNeXt V1-B      & $224^2$     & 89M         &15.4G       &85.8 \\
\gr
ConvNeXt V2-B      & $224^2$     & 89M         &15.4G       &\textbf{86.8}\textcolor{Highlight}{($+$\textbf{1.0})} \\
ConvNeXt V1-B      & $384^2$     & 89M         &45.2G       &86.8 \\
\gr
ConvNeXt V2-B      & $384^2$     & 89M         &45.2G       &\textbf{87.7}\textcolor{Highlight}{($+$\textbf{0.9})} \\
\hline
ConvNeXt V1-L      & $224^2$     & 198M         &34.4G       &86.6 \\
\gr
ConvNeXt V2-L      & $224^2$     & 198M         &34.4G       &\textbf{87.3}\textcolor{Highlight}{($+$\textbf{0.7})} \\
ConvNeXt V1-L      & $384^2$     & 198M         &101.1G       &87.5 \\
\gr
ConvNeXt V2-L      & $384^2$     & 198M         &101.1G       &\textbf{88.2}\textcolor{Highlight}{($+$\textbf{0.7})} \\
\hline
ConvNeXt V1-XL      & $224^2$     & 350M         &60.9G       &87.0 \\
ConvNeXt V1-XL      & $384^2$     & 350M         &179.0G       &87.8 \\
\hline
\gr
ConvNeXt V2-H      & $384^2$     & 660M         &337.9G       &\textbf{88.7} \\
\gr
ConvNeXt V2-H      & $512^2$     & 660M         &600.8G       &\textbf{88.9} \\
\end{tabular}
\caption{\textbf{ImageNet-22K intermediate fine-tuning results} with a single 224$\times$224 crop. The improvement over the V1 supervised model is shown in parentheses.
}
\vspace{-.5em}
\label{tab:v1_v2_comp_im22k}
\end{table}

\section{Complete comparisons with V1}
\label{sec:comp}
In Tables~\ref{tab:v1_v2_comp_im1k} and \ref{tab:v1_v2_comp_im22k}, we present detailed experiment-level comparisons between ConvNeXt V1~\cite{liu2022convnet,rw2019timm} and V2. In particular, Table \ref{tab:v1_v2_comp_im1k} shows ImageNet-1K fine-tuning results using eight models: Atto, Femto, Nano, Pico, Tiny, Base, Large, and Huge, which range from low-compute (Atto, 3.7M) to large-capacity models (Huge, 660M). We see a consistent and significant improvement across all models. The best performance is achieved when the architecture is upgraded from V1 to V2 and the self-supervised learning framework FCMAE is used, demonstrating the effectiveness of the co-design. In Table \ref{tab:v1_v2_comp_im22k}, we present ImageNet-22K intermediate fine-tuning results. The pre-training and fine-tuning process consists of three steps: 1) \convmae pre-training, 2) ImageNet-22K fine-tuning, and 3) ImageNet-1K fine-tuning. Here, we focus on five V2 models: Nano, Tiny, Base, Large and Huge.
We see consistent improvement over the V1 counterparts. In particular, the V2 Base (86.8\%/87.7\%) and Large (87.3\%/88.2\%) models outperform the next-level model sizes of V1, which are the Large (86.6\%/87.5\%) and XLarge (87.0\%/87.8\%) models. The V2 Huge model also achieves a new state-of-the-art with a performance of 88.9\%. Our proposal demonstrates that pure convolutional models can also be strong and scalable vision learners with mask-based pre-training.

\begin{figure}\centering
\vstretch{1.}{
\includegraphics[width=0.45\textwidth]{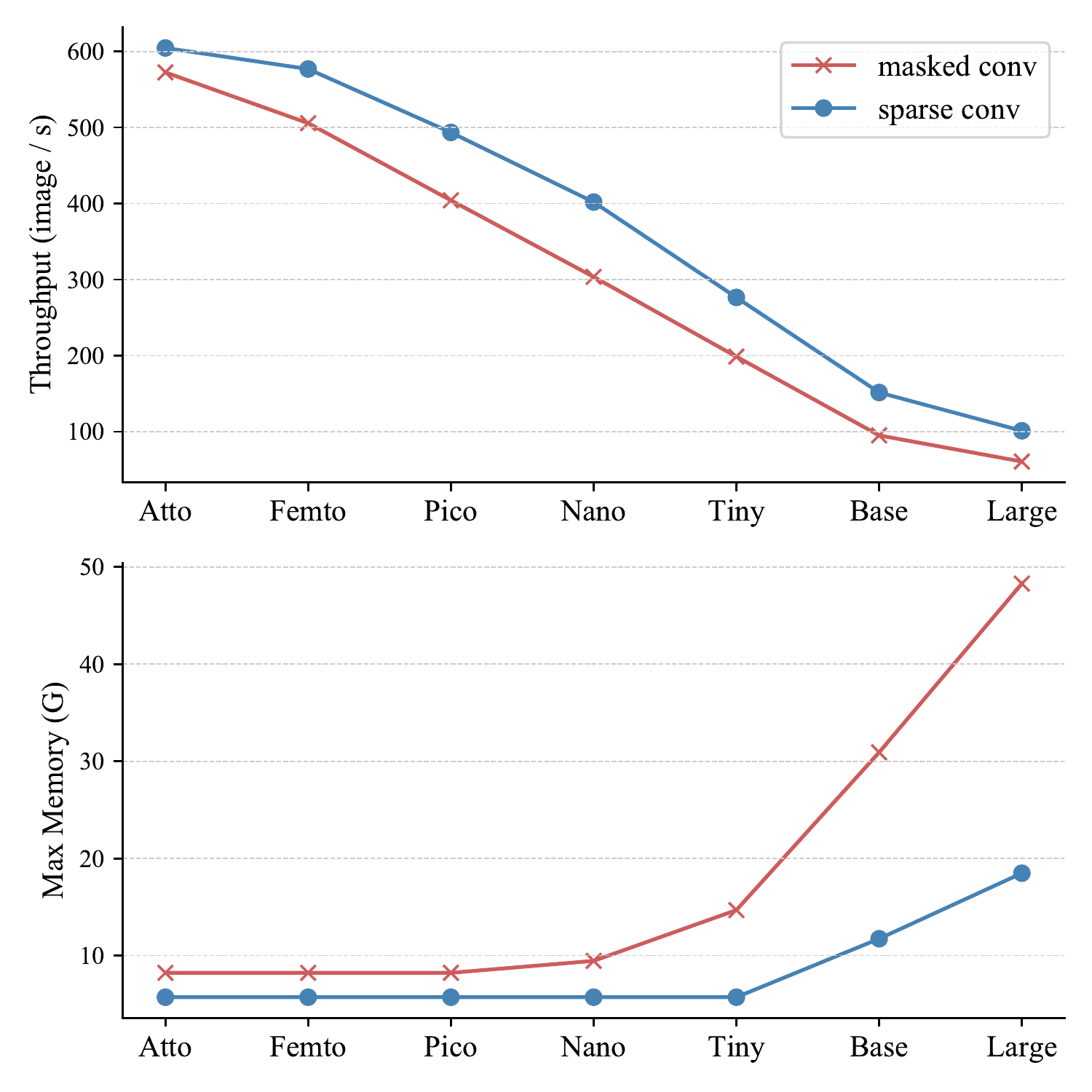}
}
\captionof{figure}{
\textbf{Sparse encoding efficiency.}
Under the pre-training setup, we measure the training throughput (image/s) and max GPU memory usage (G). The per GPU batch size is 64, and the throughput values are measured using 20 forward and backward steps. Our results show that the sparse convolution-based encoder allows for improved pre-training efficiency compared to the dense masked convolution-based counterpart.}
\label{fig:sparse_eff}
\end{figure}

\begin{figure*}\centering
\vstretch{1.}{
\includegraphics[width=1.\linewidth]{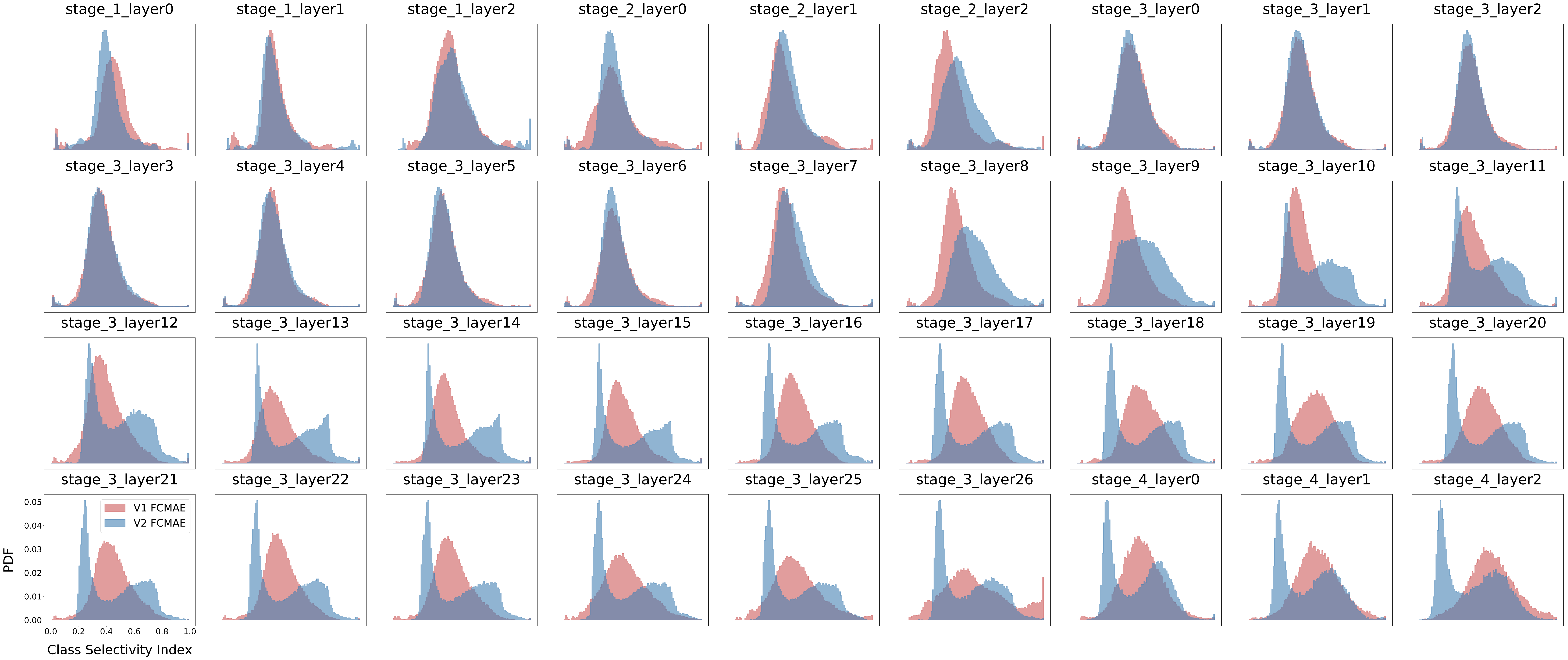}
}
\vspace{-1em}
\captionof{figure}{
\textbf{Class selectivity index distribution.} 
The x-axis and y-axis show the class selectivity index and its density (PDF), respectively. Using the ImageNet-1K validation dataset, we calculated the class selectivity index distribution of both FCMAE pre-trained ConvNeXt V1 (red) and V2 (blue). While they tend to match closely in the early stages, the distribution becomes different in the deep layers. V2 tends to include more class-generic features in the later stages.
}
\label{fig:cls_select}
\end{figure*}

\section{Further Analyses}
\label{sec:anal}
\paragraph{Sparse encoding efficiency.} One of the key design choices in our FCMAE framework is the use of sparse convolution~\cite{graham2017submanifold,graham20183d,choy20194d} during pre-training. The primary purpose is to block the flow of information from the masked region and facilitate masked autoencoder pre-training. As a byproduct, it also offers improved computational and memory efficiency during pre-training, as the kernels only apply to the visible pixels. However, we note that the sparse convolution libraries~\cite{choy20194d,spconv2022} are not highly optimized for modern hardware, and the efficiency achieved usually depends on the frameworks~\cite{Abadi2016,paszke2017automatic,jax2018github} used in practice.

To better understand the actual pre-training efficiency achieved using sparse convolution, we conducted benchmark experiments using a controlled setup with Minkowski Engine v0.5.4~\cite{choy20194d} and PyTorch~\cite{paszke2017automatic}. We simulated the pre-training masked input (image size 224$\times$224, masking ratio 0.6, mask size 32$\times$32) and compared the training throughput (image/s) and max GPU memory usage (G) between the sparse convolution-based and dense masked convolution-based encoders. While the results may vary depending on the experimental environment (we used PyTorch V1.8.0, CUDA 11.1, CuDNN 8.2, and NVIDIA RTX A6000 GPU), we observed a moderate increase in pre-training efficiency, with an average of 1.3$\times$ increase in throughput and a 2$\times$ decrease in max memory usage across the models. The gap becomes more salient as the model size increases.

\paragraph{Class Selectivity Index.} FCMAE pre-trained ConvNeXt V2 has a distinctive feature characteristic compared to V1. We conducted a class selectivity index analysis on the FCMAE pre-trained weights for ConvNeXt V1 and V2 to understand this. The class selectivity index is a metric that measures the difference between the highest class-conditional mean activity and all other class-conditional mean activities. The final normalized value lies between 0 and 1, with 1 indicating that a filter activates only for a single class and 0 indicating that the filter activates uniformly for all classes. In Figure~\ref{fig:cls_select}, we plot the class selectivity index distribution for all intermediate layers in the model, using the output of every residual block. The distribution is closely matched between V1 and V2 in the early stages, but they begin to diverge in the deep layers, such as stage 3 layer 12. As the layer becomes deeper, the plot shows that V2 (bimodal) tends to include more class-generic features than V1 (unimodal). Since class-agnostic features are more transferrable~\cite{morcos2018importance}, this leads to better fine-tuning performance in downstream tasks. We leave more explorations as a future study.
\begin{table}[t]
\centering
\tablestyle{3pt}{1.}
\begin{tabular}{cl| cc | c }
~ & ~ & \multicolumn{2}{c|}{GRN functions} & ~ \\
\multicolumn{2}{c|}{case}  & {aggregation} & {normalization} & Val acc. \\
\shline
             & \gc{base}        & -      & -       & \gc{83.7}\\
\textbf{(a)} & X                & -      & -       & 83.9     \\
\textbf{(b)} & $X*\mathcal{G}(X)$         & \cmark & -       & 83.9     \\
\textbf{(c)} & $X*\mathcal{N}(X)$        & -      & \cmark  & unstable \\
\textbf{(d)} & $X*\mathcal{N}(\mathcal{G}(X))$     & \cmark & \cmark  & 84.6     \\
\end{tabular}
\vspace{.5em}
\caption{\textbf{GRN component analysis}.
We report the fine-tuning performance after the 800 epoch FCMAE pre-training.
Here, affine parameters and residual connection are omitted for clarity.
The \gc{base} denotes the ConvNeXt V1 fine-tuning performance.
The \textbf{aggregation} and \textbf{normalization} are spatial L2-norm pooling and channel-wise divisive normalization, respectively.
Case-(a) indicates a simple baseline of channel-wise scaling and shifting (with affine parameters) without explicit feature normalization.
}
\label{tab:grn_comp_abl}
\end{table}

\section{Additional Experiments}
\label{sec:exps}

\paragraph{GRN component analysis.} The proposed Global Relation Network (GRN) consists of three steps: global feature aggregation, feature normalization, and feature calibration. The main paper demonstrates that the combination of L2-norm based aggregation and divisive normalization works well in practice. Table~\ref{tab:grn_comp_abl} verifies the individual contribution of these components using ConvNeXt V2-Base as the encoder. When either component is dropped, performance significantly decreases, and the training becomes unstable if feature normalization is not preceded by global aggregation. This supports the idea that both operations work together to make GRN effective.

\begin{figure}[t!]\centering
\includegraphics[width=0.45\textwidth]{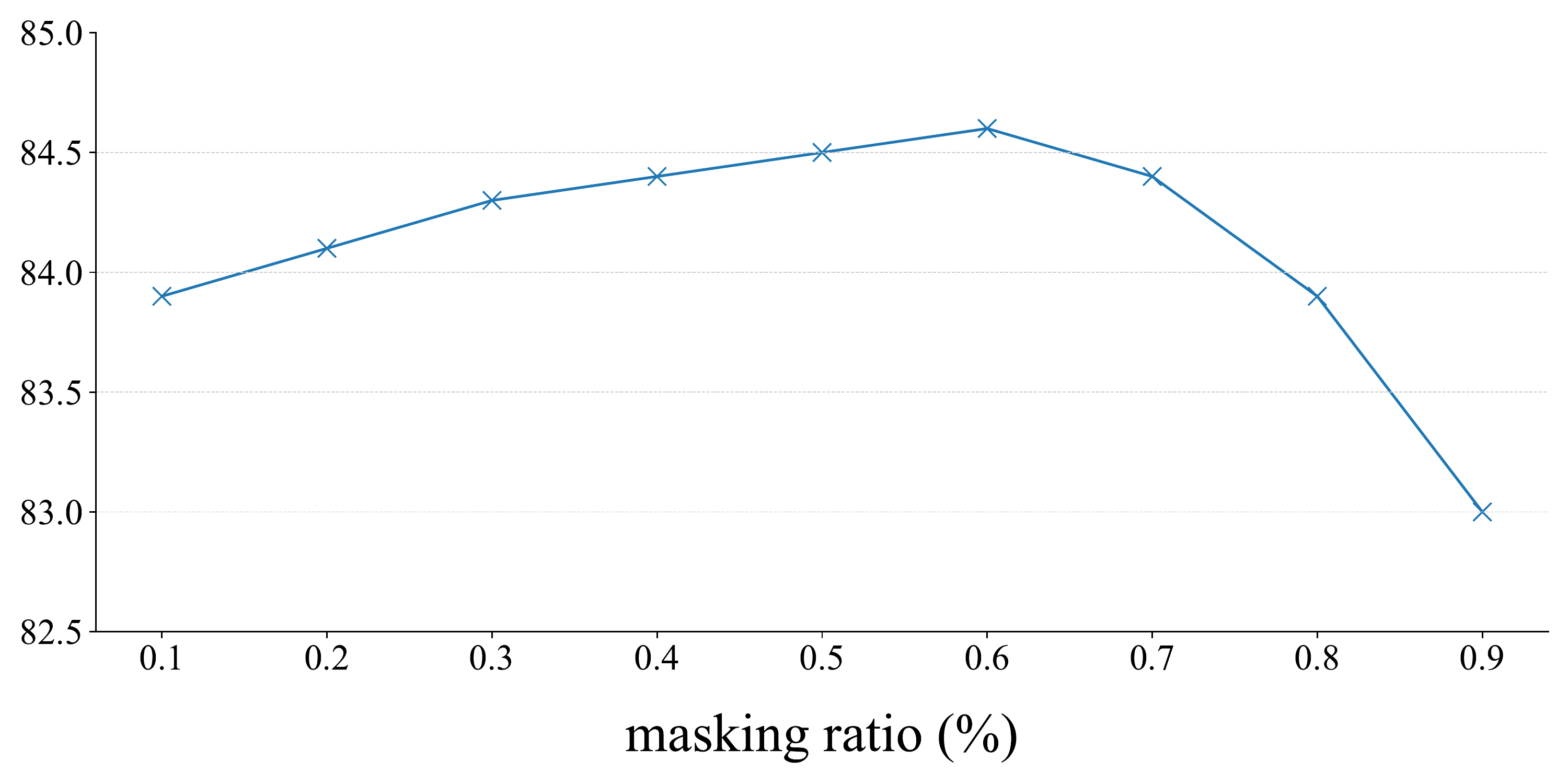}
\vspace{-.5em}
\captionof{figure}{
\textbf{Masking ratio.} We observe that a masking ratio of 0.6 provides the best result. The y-axis is ImageNet-1K accuracy (\%).
}
\label{fig:mask_ratio}
\vspace{-.5em}
\end{figure}

\paragraph{Masking ratios.} We conduct a hyper-parameter analysis on the masking ratio for a mask size of $32 \times 32$. The results, shown in Figure~\ref{fig:mask_ratio}, suggest that a masking ratio in the range of 0.5 to 0.7 produces the best results, with a masking ratio of 0.6 providing the highest performance. The model's performance declines at the two extremes of either removing or leaving 90\% of the input information, although it is more robust when more information is retained.

\paragraph{Comparison with contrastive SSL.}
In this work, we compare the performance of the two dominant self-supervised learning (SSL) approaches: contrastive learning~\cite{He2020,Chen2020,Caron2020,Grill2020,Chen2021,Caron2021,Chen2021a} and masked image modeling~\cite{Bao2021,he2022masked,xie2022simmim}. Specifically, we compare the end-to-end fine-tuning performance of MoCoV3~\cite{Chen2021a}, the current state-of-the-art contrastive learning method, with our proposed FCMAE framework using the same ConvNeXt V2-Base as the encoder. We follow the default pre-training and fine-tuning recipes for each approach and present the results below.
\begin{center}\vspace{-.2em}
\tablestyle{4pt}{1.05}
\begin{tabular}{x{60}x{60}x{60}}
Sup, 300ep. & MoCo V3 & FCMAE \\
\shline
84.3 & 83.7 & 84.9
\end{tabular}\vspace{-.2em}
\end{center}
We use the 300-epoch supervised learning baseline as a reference. The above table shows that FCMAE leads to better representation quality than MoCo V3 and also outperforms the supervised baseline. This is consistent with the recent observations that masked image modeling offers superior results over contrastive learning-based SSL for end-to-end fine-tuning. In this work, this success was also made possible with pure ConvNets.

{\small
\bibliographystyle{ieee_fullname}
\bibliography{egbib}
}

\end{document}